\documentclass[lettersize,journal]{IEEEtran}
\usepackage{amsmath,amsfonts,amssymb,amsbsy,bm,paralist,theorem,color}
\usepackage{graphicx,algorithmic,algorithm,relsize}
\usepackage{multicol}
\usepackage{multirow}
\usepackage{caption}
\usepackage{cite}
\usepackage{pifont}
\usepackage{bbm}

\usepackage[caption=false,font=footnotesize,labelfont=rm,textfont=rm]{subfig}
\usepackage{float}

\usepackage{epstopdf} 

\DeclareMathOperator*{\st}{s.t.}

\graphicspath{{fig/}}

\definecolor{orange}{RGB}{255,107,0}
\definecolor{green}{RGB}{0,160,20}

\begin{document}
	\title{Client Orchestration and Cost-Efficient Joint Optimization for NOMA-Enabled Hierarchical Federated Learning}
	\author{Bibo~Wu,~\IEEEmembership{}
		Fang~Fang,~\IEEEmembership{Senior Member, IEEE,}
		Xianbin Wang,~\IEEEmembership{Fellow, IEEE,}
		Donghong Cai,~\IEEEmembership{Member, IEEE,}\\
		Shu Fu,~\IEEEmembership{Member, IEEE,}
		and Zhiguo Ding,~\IEEEmembership{Fellow, IEEE}
		
		\thanks{Bibo Wu, Fang Fang and Xianbin Wang are with the Department of Electrical and Computer Engineering, and Fang Fang is also with the Department of Computer Science, Western University, London, ON N6A 3K7, Canada (e-mail: \{bwu293, fang.fang, xianbin.wang\}@uwo.ca).}
		\thanks{Donghong Cai is with the College of Cyber Security, Jinan University, Guangzhou, 510632, China (e-mail: dhcai@jnu.edu.cn).}
		\thanks{Shu Fu is with the Department of Microelectronics and Communication Engineering, Chongqing University, Chongqing, 400044, China (e-mail: shufu@cqu.edu.cn).}
		\thanks{Zhiguo Ding is with the Department of Electrical Engineering and Computer Science, Khalifa University, Abu Dhabi, UAE, and the Department of Electrical and Electronic Engineering, University of Manchester, Manchester, M13 9PL, UK (e-mail: zhiguo.ding@manchester.ac.uk).}
		
		}
	
 	\maketitle
	\begin{abstract}
	Hierarchical federated learning (HFL) shows great advantages over conventional two-layer federated learning (FL) in reducing network overhead and interaction latency while still retaining the data privacy of distributed FL clients.
	However, the communication and energy overhead still pose a bottleneck for HFL performance, especially as the number of clients raises dramatically.
	To tackle this issue, we propose a non-orthogonal multiple access (NOMA) enabled HFL system under semi-synchronous cloud model aggregation in this paper, aiming to minimize the total cost of time and energy at each HFL global round.
	Specifically, we first propose a novel fuzzy logic based client orchestration policy considering client heterogenerity in multiple aspects, including channel quality, data quantity and model staleness.
	Subsequently, given the fuzzy based client-edge association, a joint edge server scheduling and resource allocation problem is formulated. 
	Utilizing problem decomposition, we firstly derive the closed-form solution for the edge server scheduling subproblem via the penalty dual decomposition (PDD) method.
	Next, a deep deterministic policy gradient (DDPG) based algorithm is proposed to tackle the resource allocation subproblem considering time-varying environments.
	Finally, extensive simulations demonstrate that the proposed scheme outperforms the considered benchmarks regarding HFL performance improvement and total cost reduction.
		
	\end{abstract}
	

	\section{Introduction}
	\IEEEPARstart{T}{he} burgeoning Internet of Things (IoT) has greatly increased the number of intelligent devices and applications, which leads to the humongous amount of data traffic.
	To this end, machine learning (ML) becomes necessary due to its great potential in dealing with the extensive amount of data \cite{FLIoT}.
	In general, the traditional ML models are trained using a centralized way, which requires a central server to assemble the distributed raw data of users for model training \cite{FLmag}.
	This centralized paradigm inevitably raises the data privacy concern for users.
	In response, a decentralized ML framework called federated learning (FL) was proposed by Google and has been investigated extensively for the past few years \cite{FL_Google}.
	Under the framework of FL, the edge entities, termed clients, are allowed to train a global model collaboratively with local trained model parameters rather than their local private data.
	As a direct benefit, the global FL model can be well trained while preserving the data privacy.
	
	The implementation of conventional FL normally based on the two-layer framework, which can be divided into two classes, including the client-cloud FL \cite{DLmag} and the client-edge FL \cite{FL_edge}.
	However, the heavy communication overhead and the inferior FL performance are the two major drawbacks for these two types of FL, respectively.
	In response, the hierarchical federated learning (HFL) has been proposed recently, which exploits the client-edge-cloud three-layer structure to realize better performance \cite{HFL_ICC}.
	In HFL, each edge server plays the role of intermediate node which performs the edge model aggregation for local models uploaded by the clients in vicinity. 
	Next, the aggregated edge models are transmitted by edge servers to the cloud server for global model aggregation when the accuracy threshold for edge model training is satisfied.
	As a result, HFL significantly reduces the transmission latency and energy overhead between clients and the remote cloud server.
	Besides, the training performance can be guaranteed by the cloud server which is able to orchestra sufficient clients via edge servers.

	Nevertheless, the communication latency and energy overhead become the bottleneck for the further improvement of HFL performance as the number of clients raises.
	Given that the densely distributed clients normally can communicate with more than one edge server, the client orchestration, i.e., the association between clients and edge servers, plays a vital role in HFL systems. 
	Through effective design of client-edge association, more efficient resource allocation among clients can be achieved, leading to improved client-edge communication and further reduction of HFL training time and energy overhead.
	Moreover, using non-orthogonal resource-efficient channel access is also helpful in enhancing HFL outcome by reducing access latency and increasing network throughput when communication resources become scarce.
	In particular, non-orthogonal multiple access (NOMA) \cite{Ding_Survey_2017} allows multiple clients to transmit their local model parameters simultaneously on the same channel, which has been regarded as a significant enabler to enhance communication efficiency in HFL systems.
	Therefore, the investigation towards client orchestration and efficient resource allocation in HFL over NOMA networks deserves to be conducted.

	\subsection{Related Literature}
	In recent years, abundant of research has been conducted to improve the performance of two-layer FL and HFL.

	\subsubsection{Two-layer FL performance improvement}
	Considering the resource bottleneck in the two-layer FL systems, a variety of mechanisms have been proposed to improve FL performance in existing works, including model compression \cite{Prun, Compression1}, client scheduling \cite{FLCS3, FLCS4, FL_BBW} and resource allocation \cite{FLT3, FLT4,FLT1, FLT2, FLloss1}.
	Specifically, the authors in \cite{Prun} applied network pruning to maximize the FL convergence rate, while the gradient compression method was utilized in \cite{Compression1} to realize the tradeoff between FL convergence rate and latency.
	With the consideration of resource scarceness, client selection is crucial to improve FL performance.
	In \cite{FLCS3}, a client orchestration policy based on probability was designed by minimizing the total communication time.
	The concepts of reputation and contract were applied to devise the client selection in \cite{FLCS4}, where the trustworthy clients are selected.
	Recently, the authors in \cite{FL_BBW} considered an emerging criterion, i.e., model staleness, for client scheduling in FL systems, and proposed an age-based selecting policy to achieve better FL performance.
	Resource allocation is also vital for the implementation of communication-efficient FL. 
	To minimize the convergence time of FL, the joint resource allocation and client selection problem was formulated to optimize transmitting power \cite{FLT3}, computing frequency \cite{FLT4} and uplink resource block \cite{FLT1}.
	Some other works paid attention to the minimization of global FL loss with efficient resource allocation \cite{FLT2, FLloss1}.
	However, the aforementioned two-layer FL works are not suitable for the large-scale systems, owing to the long training latency and heavy energy overhead.
		
	\subsubsection{HFL design and analysis}	
	By taking advantage of the conventional two-layer FL, the HFL with the client-edge-cloud structure was proposed and studied extensively \cite{xu2021adaptive, wen2022joint, HFEL, liu2022joint, liu2022hierarchical}.
	Specifically, considering the edge aggregation interval control, \cite{xu2021adaptive} investigated the multi-objective optimization problem of global loss and training latency in a HFL system.
	Differently, the convergence bound for one HFL round was analyzed in \cite{wen2022joint}, and the joint resource allocation and edge server scheduling problem was investigated.
	In \cite{HFEL} and \cite{liu2022joint}, the client-edge association scheme was carefully designed to achieve efficient resource allocation, while \cite{liu2022joint} considered the heterogeneous data distribution among clients.
	The authors in \cite{liu2022hierarchical} took model quantization into account to analyze the HFL convergence bound, which was then applied to the design of edge and cloud model aggregation interval.
	
	To achieve a further improvement of training performance and communication efficiency, several promising technologies have been considered in HFL systems, including  NOMA \cite{HFL_DDPGNOMA}, blockchain \cite{HFL_chain} and deep reinforcement learning (DRL) \cite{HFL_DQN2, HFL_multiDDPG}.
	More specifically, \cite{HFL_DDPGNOMA} applied NOMA in HFL systems and formulated a joint problem regarding client orchestrarion and resource allocation.
	Considering the willingness of clients to transmit their local models during HFL training, \cite{HFL_chain} proposed a blockchain-based incentive scheme to achieve better performance.
	The rapid development of DRL makes the integration of DRL with HFL more prevalent.
	A two-layer deep Q-leaning based incentive mechanism was proposed in \cite{HFL_DQN2} to motivate clients for aggregation, which contributes to the high-quality HFL model training.
	The authors in \cite{HFL_multiDDPG} proposed a multi-agent deep deterministic policy gradient (DDPG) based algorithm to minimize the HFL global loss.
	Nevertheless, none of the above HFL works touched upon the client-edge association under multiple criteria, hence failing to balance the client heterogenerity, which limits them in practical wireless scenarios.

	\subsection{Motivations and Contributions}
	From above observations, HFL shows great superiority than traditional two-layer FL in reducing training latency and energy cost.
	Nevertheless, there is still a bottleneck of communication and energy overhead for the further improvement of HFL performance when deploying in large-scale systems.
	Intelligent client orchestration is one of the solutions to address this issue.
	However, the client-edge association schemes in the HFL literature only depend on a single criterion and fail to consider the client heterogeneity in multiple aspects, e.g., data, model and communication quality, which makes them unsuitable in practical wireless networks.
	Hence, it is essential to design multi-criteria based client-edge association schemes for realizing communication-efficient HFL systems.
	Besides, NOMA shows great potential in addressing the resource scarceness issue via improved access and communication efficiency, which is beneficial to facilitate HFL convergence.
	Although \cite{HFL_DDPGNOMA} has conducted the research on NOMA-enabled HFL system, the impact of NOMA on HFL performance was not analyzed.
	
	Motivated by above analysis, we propose a novel NOMA enabled HFL system in this paper, and design a multi-criteria client orchestration policy based on fuzzy logic. 
	We investigate the joint optimization problem of edge server scheduling and resource allocation to minimize the total time and energy cost.
	The closed-form solution of edge server scheduling is obtained through the penalty dual decomposition (PDD) approach, and a DDPG-based algorithm is introduced to allocate resources efficiently. 
	The main contributions of this paper are summarized as follows:
	\begin{enumerate}[1)]
		\item We propose a NOMA enabled HFL system for reduced communication and energy overhead, where the semi-synchronous cloud aggregation is considered.
		The implementation of HFL training is divided into the client-edge phase and the edge-cloud phase.
		The corresponding time consumption and energy cost are formulated at each phase.

		\item We devise a multi-criteria client-edge association policy according to fuzzy logic in light of client heterogenerity, including wireless channel quality, data quantity as well as model staleness.
		The membership functions for fuzzy inputs and fuzzy outputs, and the fuzzy rules are devised carefully.
		The final defuzzied output of each client implies its competency level to associate with the edge server, which is utilized to guide the client orchestration.

		\item Aiming to minimize the total cost at one HFL round, a joint edge server scheduling and resource allocation problem is formulated, which is a non-convex mixed-integer non-linear programming (MINLP) problem.
		In order to solve it effectively, the original problem is decomposed into two subproblems for edge server scheduling and resource allocation, respectively.
		A PDD based algorithm with the double-loop structure is proposed to derive the closed-form solution for edge server scheduling.
		We resort the DRL method for resource allocation with the consideration of time-varying channels, and develop a DDPG-based algorithm to address the continuous resource allocation subproblem.
		
	\end{enumerate}

	\subsection{Organization}
	We organize the remainder of this paper as follows.
	Section II presents the NOMA enabled HFL system model.
	In Section III, the fuzzy based client orchestration scheme is designed.
	In Section IV, we propose the edge scheduling and resource allocation algorithms to achieve total cost minimization.
	Extensive simulation results are shown in Section V, and we conclude this paper in Section VI.

	\section{System Model}
	
	\begin{figure}[!t]
		\centering
		\includegraphics[width=0.48\textwidth]{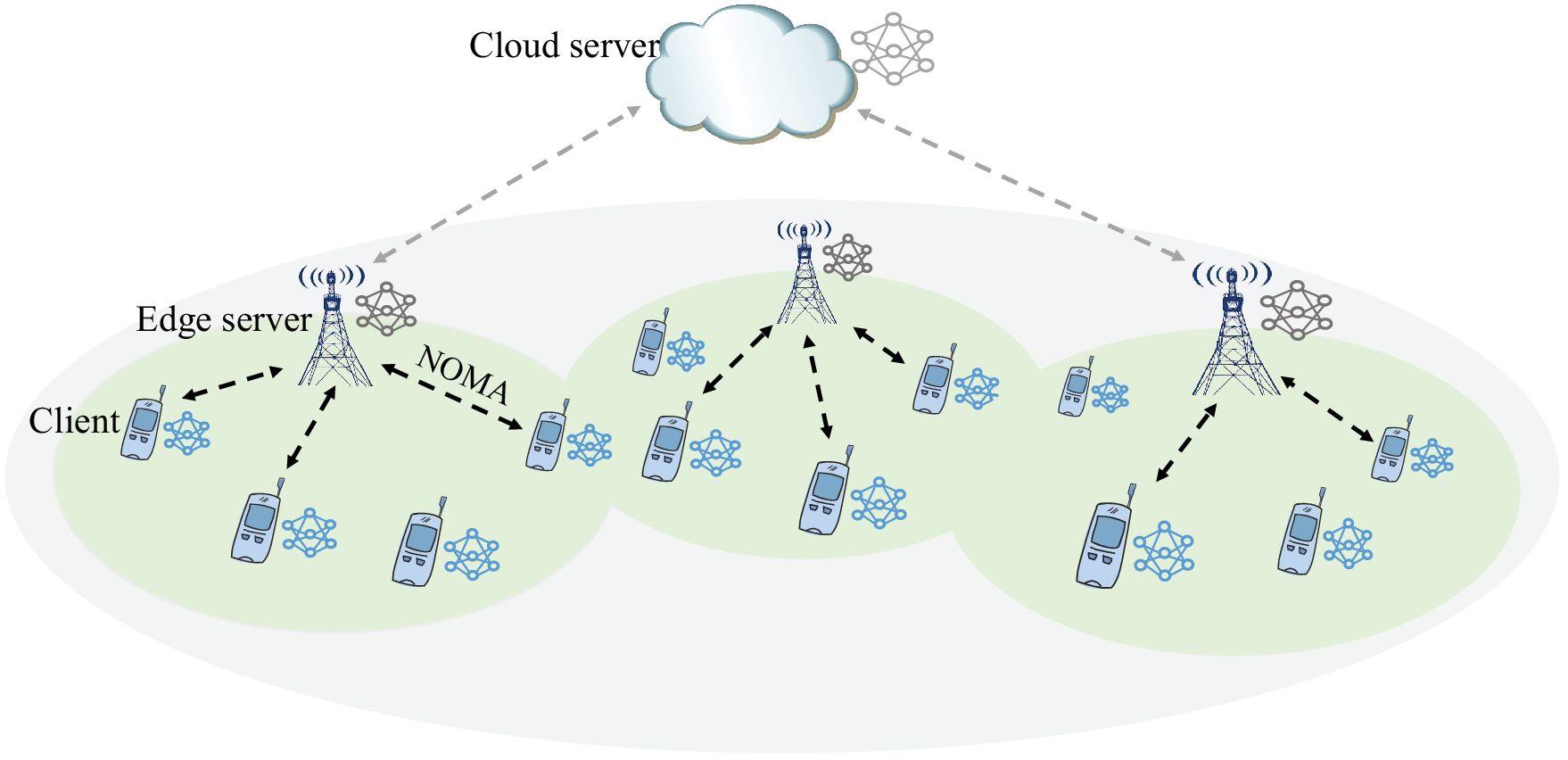}
		\caption{NOMA enabled HFL system.}
		\label{system model}
	\end{figure}

	Fig. \ref{system model} illustrates the proposed HFL system, where a remote cloud server coordinates $M$ edge servers denoting by ${\cal{M}} \in \left\{ {1,2, \ldots ,M} \right\}$ and each edge server $m$ associates with a set of clients ${\cal{N}}_m \in \left\{ {1,2, \ldots ,N_m} \right\}$.
	Let ${\cal{N}} \in \left\{ {1,2, \ldots ,N} \right\}$ denotes the client set, and we assume $N \gg M $ in the system.
	For client $n$, it owns a local training data set denoted as ${{\cal{D}} _n} = \left\{ {\left( {{{\bf{x}}_i},{y_i}} \right)} \right\}_{i = 1}^{ {{D_n}} }$, in which ${D_n}$ denotes the total number of data samples, ${{\bf{x}}_i}$ and ${y_i}$ are the 
	$i$-th data sample and the corresponding label, respectively.
	We divide the HFL implementation into two phases, including the client-edge phase as well as the edge-cloud phase, and the details are presented as follows.
	
	
	\subsection{Client-Edge Phase}
	\subsubsection{Local model training} At this stage, client $n$ trains the local model utilizing its local data samples iteratively until reaching the local training accuracy $\theta \in (0, 1)$.
	We assume that the local training accuracy is the same among all clients in the HFL system.
	Denote client $n$'s local loss function by ${l_n}\left( {{\bf{w}}_n,{{\bf{x}}_i},{y_i}} \right)$, where ${\bf{w}}_n$ is the model paramater which characterizes the output label ${y_i}$ with the input data sample ${{\bf{x}}_i}$.
	Based on this, we define the local loss of client $n$ as 
	\begin{equation}\label{}
		\begin{aligned}
			{L_n}\left( {\bf{w}}_n \right) = \frac{1}{{{D_n}}}\sum\limits_{i = 1}^{{D_n}} {{l_n}\left( {{\bf{w}}_n,{{\bf{x}}_i},{y_i}} \right)}.
		\end{aligned}
	\end{equation}
	At the $t$-th local iteration, client $n$ updates its local FL model in the following gradient descent manner:
	\begin{equation}\label{}
		\begin{aligned}
			{\bf{w}}_n^t = {\bf{w}}_n^{t-1} - \eta \nabla {L_n}\left( {{\bf{w}}_n^{t - 1}} \right),
		\end{aligned}
	\end{equation}
	in which $ \eta$ denotes the learning rate.
	To satisfy local training accuracy $\theta$, client $n$ is supposed to run the local model update many times until satisfing $\left\| {\nabla {L_n}\left( {{\bf{w}}_n^t} \right)} \right\| \le \theta \left\| {\nabla {L_n}\left( {{\bf{w}}_n^{t - 1}} \right)} \right\|$.
	The local iteration number is given by \cite{LocalAccu1}
	\begin{equation}\label{}
		\begin{aligned}
			 {\tau _1} = \mu \log \left( {\frac{1}{\theta }} \right),
		\end{aligned}
	\end{equation}
	wherein $\mu$ represents the learning task related constant.
	
	For any client $n$, it utilizes the equipped CPU resources to finish the local model training task.
	Define ${f_n}$ as the CPU frequency of client $n$.
	We can calculate its local training time with ${\tau _1}$ local iterations by
	\begin{equation}\label{}
		\begin{aligned}
			t_n^{cmp} = \frac{{\tau _1}{{c_n}{D_n}}}{{{f_n}}},
		\end{aligned}
	\end{equation}
	where ${c_n}$ represents the required number of CPU cycles to deal with one data sample for client $n$.
	The corresponding energy consumption during the entire local model training process can be given as \cite{LocComEn}
	\begin{equation}\label{}
		\begin{aligned}
			e_n^{cmp} = {\tau _1}\frac{{{\beta _n}}}{2}f_n^2{c_n}{D_n},
		\end{aligned}
	\end{equation}
	wherein $\frac{{{\beta _n}}}{2}$ is the computing chipsets' effective capacitance coefficient at client $n$.

	\subsubsection{Client-edge transmission} In this paper, we propose the utilization of NOMA transmission between clients and the associated edge server.
	NOMA enables multiple clients to take up the same channel to simultaneously upload their local models, hence improving the communication access and efficiency.
	For the sake of simplicity, we omit the iteration index in the following content.
	Denote $s_n$ by the transmitting symbol of client $n$, which can be normalized as $\left\| {s_n} \right\|_2^2 = 1$.
	We define $p_n$ as the client $n$'s transmitting power.
	Let $h_{n,m}$ be the gain of wireless channel between client $n$ and its associated edge server $m$.
	Note that the association between clients and edge servers is a vital issue to be carefully considered in HFL, as a client has the access to associate with more than one edge server but only associates with one specific edge server at each round.
	With NOMA transmission, the received signal at edge server $m$ is the superposition form of its orchestrated clients' signals, which is expressed as 
	\begin{equation}\label{}
		\begin{aligned}
			{y_m} = \sum\limits_{n = 1}^{N_m} {\left( {\sqrt {{p_n}} {h_{n,m}}{s_n}} \right)}  + {n_m},
		\end{aligned}
	\end{equation}
	wherein ${n_m} \sim {\cal{CN}} \left( {0,{\sigma_m ^2}} \right)$ denotes the additive white Gaussian noise (AWGN) whose mean is 0 and variance is ${\sigma_m ^2}$.
	
	In NOMA, the interference among clients is introduced due to transmitting simultaneously over the same channel.
	To tackle this issue, the successive interference cancellation (SIC) approach has been commonly adopted in NOMA-enabled systems \cite{RIS_NOMA_BBW}.
	Particularly, for the uplink NOMA transmission, by regarding other signals as interference, the receiver decodes the most powerful signal first.
	After subtracting the decoded signal from the superpositioned signal, the receiver subsequently performs decoding to the second strongest signal in the same manner.
	This approach proceeds until decoding all signals successfully.
	Note that the design of an optimal SIC decoding order is beyond the scope of this paper.
	Thus, for simplicity of analysis, we make assumption that the decode order is decided based on the received power, i.e. $\sqrt {{p_1}} {\left| {{h_{1,m}}} \right|^2} > \sqrt {{p_2}} {\left| {{h_{2,m}}} \right|^2} >  \cdots  > \sqrt {{p_{{N_m}}}} {\left| {{h_{{N_m},m}}} \right|^2}$, which indicates that for the edge server $m$, the client 1's signal is the first decoded signal, while the client ${N_m}$'s signal is the last decoded one.
	In this case, the signal-to-interference-to-noise ratio (SINR) of client $n$ at the edge server $m$ can be presented as 
	\begin{equation}\label{}
		\begin{aligned}
			\text{SINR}_{n,m} = \frac{{{p_n}{{\left| {{h_{n,m}}} \right|}^2}}}{{\sum\limits_{j = n + 1}^{{N_m}} {{p_j}{{\left| {{h_{j,m}}} \right|}^2}}  + \sigma _m^2}},
		\end{aligned}
	\end{equation}
	where $\sum\limits_{j = n + 1}^{{N_m}} {{p_j}{{\left| {{h_{j,m}}} \right|}^2}}$ denotes the interference caused by the clients which are decoded after client $n$.
	In particular, the SINR for client ${N_m}$ is calculated by $ \text{SINR}_{{N_m},m} = \frac{{{p_{N_m}}{{\left| {{h_{{N_m},m}}} \right|}^2}}}{{ \sigma _m^2}}$.
	Based on the derived SINR, client $n$'s achievable data rate at the edge server $m$ is expressed as
	\begin{equation}\label{}
		\begin{aligned}
			{R_{n,m}} = B{\log _2}\left( {1 + \text{SINR}_{n,m}} \right),
		\end{aligned}
	\end{equation}
	wherein $B$ denotes the channel bandwidth.
	
	Given the client $n$'s achievable data rate, its NOMA transmission time at one edge iteration is given by 
	\begin{equation}\label{}
		\begin{aligned}
			t_{n,m}^{com} = \frac{{{d_n}}}{{{R_{n,m}}}},
		\end{aligned}
	\end{equation}
	where $d_n$ denotes the size of client $n$'s local model parameter, which is considered to be the same among all clients because of the similar numbers of elements in local models \cite{FLT1}.
	The corresponding energy consumption can be calculated as 
	\begin{equation}\label{}
		\begin{aligned}
			e_{n,m}^{com} = {p_n} {t_{n,m}^{com}}.
		\end{aligned}
	\end{equation}
	
	\subsubsection{Edge model aggregation} At this stage, for edge server $m$, it would aggregate the local FL model parameters sent by its orchestrated clients set ${\cal{N}}_m$ in an averaging manner, which is defined as
	\begin{equation}\label{}
		\begin{aligned}
			{{\bf{w}}_m} = \frac{{\sum\limits_{n \in {{\cal N}_m}} {{D_n}{{\bf{w}}_n}} }}{D_{{{\cal N}_m}}}, 
		\end{aligned}
	\end{equation}
    where ${D_{{{\cal N}_m}}} = {\cup_{n \in {{\cal N}_m}}{{D_n}} }$ denotes the size of edge server $m$'s aggregated dataset.
	Then, each edge server broadcasts the aggregated edge model parameter to its orchestrated clients to perform the local model training of the next iteration.
	This process is conducted iteratively by the edge server until meeting the edge accuracy requirement $\xi $ which is the same for all edge servers.
	The edge iteration number can be calculated by \cite{EdgeAcc}
	\begin{equation}\label{}
		\begin{aligned}
			{\tau _2} = \frac{{\delta \left( {\log \left( {\frac{1}{\xi }} \right)} \right)}}{{1 - \theta }},
		\end{aligned}
	\end{equation}
	in which $\delta$ denotes the constant regarding learning tasks.
	
	Considering there are abundant resources for edge servers to conduct the edge model aggregation and broadcasting, the corresponding time and energy consumption are reasonably ignored in this paper.
	Hence, after ${\tau _2}$ edge iterations, the total time and energy consumption for client set ${\cal{N}}_m$ orchestrated by the edge server $m$ can respectively be expressed as 
	\begin{equation}\label{TimeEdge}
		\begin{aligned}
			T_{{{\cal{N}}_m}}^{edge} = \mathop {\max }\limits_{n \in {{\cal{N}}_m}} \left\{ {\tau _2} \left( {t_n^{cmp} + t_{n,m}^{com}}\right)  \right\},
		\end{aligned}
	\end{equation}
	\begin{equation}\label{}
		\begin{aligned}
			E_{{{\cal{N}}_m}}^{edge} = \sum\limits_{n \in {{\cal{N}}_m}} {{\tau _2}\left( {e_n^{cmp} + e_{n,m}^{com}} \right)} .
		\end{aligned}
	\end{equation}
	Note that the Max in \eqref{TimeEdge} implies the edge model aggregation is synchronous, i.e., the last client to complete the model parameter uploading determines the total time consumption.
	The synchronous manner for the edge model aggregation is reasonable as each edge server's coverage is limited, leading to the moderate time consumption for each client.
	In this case, the straggler problem will not degrade HFL performance severely.

	\subsection{Edge-Cloud Phase}
	
	\subsubsection{Edge-cloud transmission} After reaching the edge accuracy, the aggregated edge model parameter is uploaded by each edge server to the cloud for global model aggregation.
	As compared to the client-edge phase, the edge-cloud phase has a much lower communication burden, we adopt the orthogonal frequency division multiple access (OFDMA) transmission at this stage.
	Note that the edge servers have relatively abundant resources for computing and communicating compared with clients.
	Thus, the corresponding resource allocation issue is not the focus of this paper.
	For ease of analysis, we omit the calculation of achievable data rate for edge servers.
	Define $p_m$ and $R_m$ as the uplink transmitting power and data rate of edge server $m$, respectively.
	As a result, its time and energy consumption for model parameter transmission can be presented by
	\begin{equation}\label{}
		\begin{aligned}
			T_m^{cloud} = \frac{{{d_m}}}{{{R_m}}},
		\end{aligned}
	\end{equation}
	\begin{equation}\label{}
		\begin{aligned}
			E_m^{cloud} = {p_m}T_m^{cloud},
		\end{aligned}
	\end{equation}
	wherein ${d_m}$ is the size of aggregated edge model parameter.

	\begin{figure}[t]
		\includegraphics[width=0.48\textwidth]{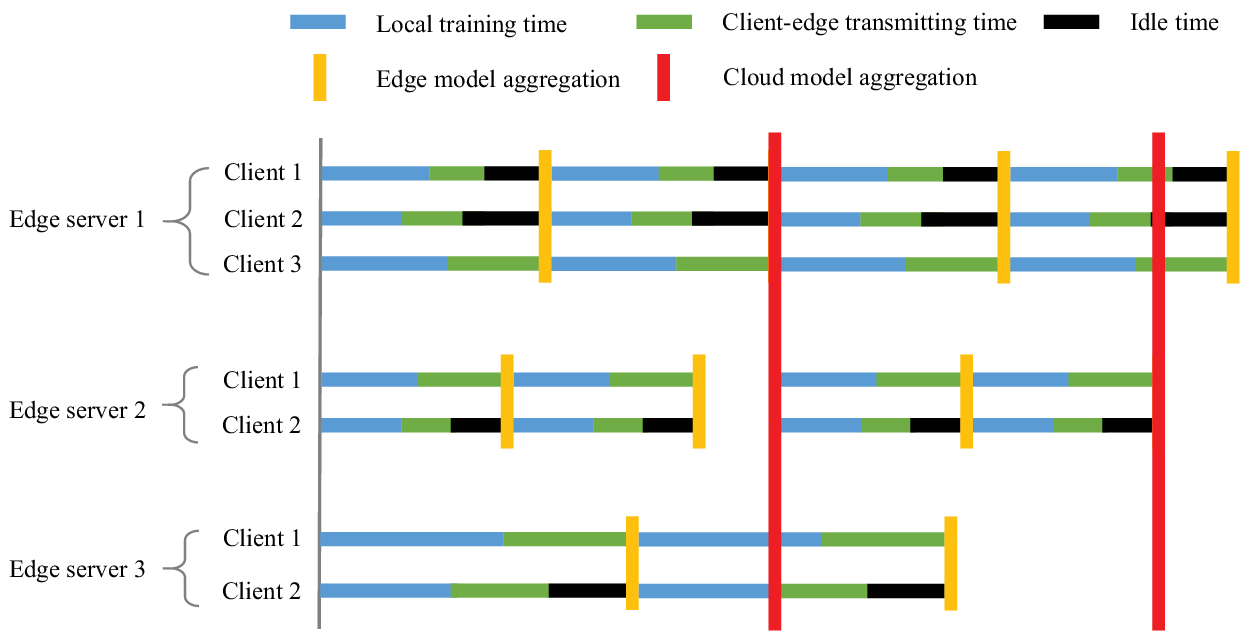}
		\centering
		\caption{Illustration of semi-synchronous cloud  aggregation.}
		\label{SemiModel}
	\end{figure}
	
	\subsubsection{Cloud model aggregation} We consider a semi-synchronous cloud model aggregation mechanism in this paper, due to the following two aspects.
	On the one hand, if we adopt the absolute synchronous cloud aggregation, the long transmission delay may cause severe straggler problem, which will degrade the HFL performance.
	On the other hand, if the absolute asynchronous cloud aggregation manner is adopted,  the cloud server would frequently communicate with the edge servers with better computing capability and communicating conditions, which may not only lead to heavy communication overhead but also lead to the overfitting risk for the global HFL model.
	The semi-synchronous cloud aggregation offers a balanced tradeoff to address these issues.
	As shown in Fig. \ref{SemiModel}, in our proposed semi-synchronous cloud aggregation, the first $M_c$ edge servers are selected by the cloud server according to their total time consumption, while the rest of edge servers will proceed their edge training and await the next cloud aggregation when they have finished the edge model uploading.
	The selected edge servers set is denoted by ${\cal{M}}_c \in \left\{ {1,2, \ldots ,M_c} \right\}$.
	Denote $z_m$ by the selecting indicator for edge server $m$, i.e., if the edge server $m$ is selected for global model aggregation, $z_m = 1$; otherwise $z_m = 0$.
	The semi-synchronous aggregation at the cloud server is given by
	\begin{equation}\label{}
		\begin{aligned}
			{\bf{w}} = \frac{{\sum\limits_{m \in {\cal M}} {{z_m}{D_{{{\cal N}_m}}}} {{\bf{w}}_m}}}{{\cup_{m \in {\cal M}} {{z_m}{D_{{{\cal N}_m}}}} }}.
		\end{aligned}
	\end{equation}
	
	Note that we ignore the time and energy cost of the cloud server for cloud model aggregation and broadcasting in this paper, as it is supposed to be powerful enough in computing and communicating.
	Therefore, we can derive the system-wide total time and energy consumption for the proposed HFL system in one global iteration, which are given by
	\begin{equation}\label{}
		\begin{aligned}
			T = \mathop {\max }\limits_{m \in {\cal M}}{z_m} \left\{ {T_m^{cloud} + T_{{{\cal N}_m}}^{edge}} \right\},
		\end{aligned}
	\end{equation}
	\begin{equation}\label{}
		\begin{aligned}
			E = \sum\limits_{m \in {\cal M}} {z_m}{\left( {E_m^{cloud} + E_{{{\cal N}_m}}^{edge}} \right)}.
		\end{aligned}
	\end{equation}

	\section{Fuzzy logic based client-edge association}
\begin{figure}[t]
		\includegraphics[width=0.48\textwidth]{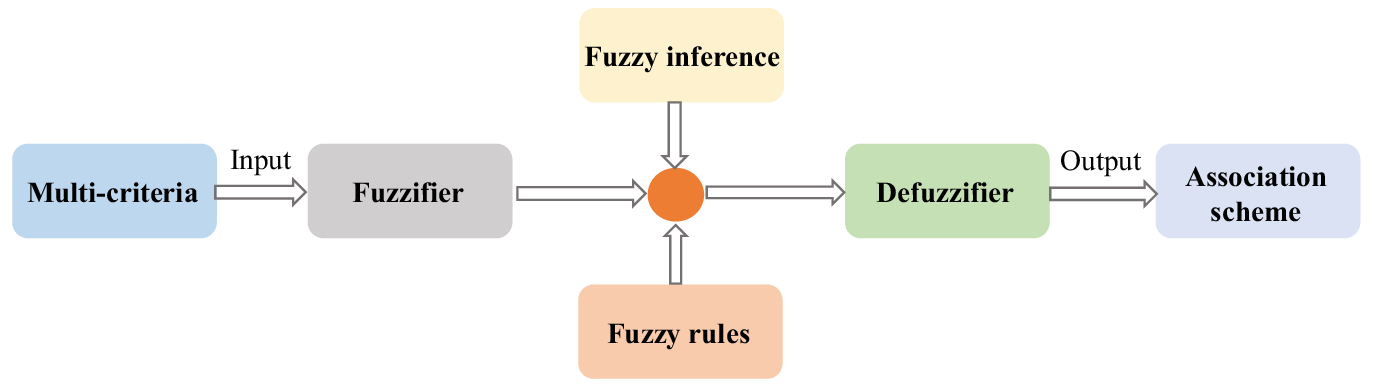}
		\centering
		\caption{Fuzzy logic based client-edge association.}
		\label{FuzzyModel}
	\end{figure}

	The association between clients and edge servers is designed in this section, which is vital in enhancing the HFL performance, especially in large-scale systems.
	Note that the number of clients far exceeds that of edge servers in our proposed HFL system.
	In this case, only a subset of clients can associate with the edge server because of the limited communication resources.
	Besides, we consider that each client can be orchestrated by different edge servers, but it can only associate with one edge server during one global iteration.
	Therefore, it requires us to carefully design the client-edge association to achieve better HFL performance.
	Previous client-edge association schemes mainly depend on a single factor or criterion, such as the distance between the edge server and clients, the data distribution of clients and so forth.
	However, such single-criterion based client-edge association schemes fail to simultaneously balance the client heterogenerity in the system, which inevitably degrades the HFL performance.
	Differently, in this paper, a multi-criteria based client-edge assocation scheme according to the fuzzy logic \cite{fuzzybook} is proposed, as shown in Fig. \ref{FuzzyModel}, by simultaneously considering the client-edge channel quality, the data quantity of client and the local model staleness.
	The details of the fuzzy logic based client-edge association scheme are presented as follows.
	
	\subsection{Fuzzy Input}
	
	\subsubsection{Channel quality (CQ)} This input variable indicates the client-edge communication environment.
	In our proposed HFL system, the client with poor CQ is not supposed to transmit its local model for edge aggregation, because it will cause long transmission delay under the synchronous model aggregation manner, which could degrade the HFL performance.
	Thus, CQ is one of the key factors needed to be considered when designing the client-edge association scheme.
	Note that in wireless communications, the CQ is affected by various factors.
	In this case, the CQ of a client in HFL cannot be determined merely by the distance from the client to the edge server.
	In other words, even if the client is located closely to the edge server, the CQ between them may still be unsatisfied due to the fading effect.
	In this case, we are supposed to choose CQ rather than the distance as a factor for the client-edge association design. 
	
	\subsubsection{Data quantity (DQ)} To improve HFL training performance, it is essential to select as many clients as possible, especially considering heterogeneous data distribution among clients. 
	In this case, sufficient data samples can be utilized to train the global HFL model.
	Nevertheless, we consider that the number of clients far exceeds that of edge servers in the proposed system, leading to few selected clients at each global round.
	Thus, to achieve better HFL performance, we need to consider how we can choose the clients with more data samples.
	Hence, DQ is another essential factor needed to be considered when designing the association scheme between the clients and the edge servers.
	
	\subsubsection{Model staleness (MS)} The MS describes the elapsed time from the client's latest selection by the associated edge server.
	Specifically, if a client has not been orchestrated by the edge server for a long time, its local model will be more stale, which is not beneficial for the global  model training and convergence \cite{AoI}.
	To realize better HFL performance, the edge servers tend to obtain local models updated timely.
	We define the client $n$'s MS value at the $i$-th global round as $A_n^i$, which can be calculated by 
	\begin{equation}\label{AoU}
		\begin{aligned}
			A_n^i = \left\{ {\begin{array}{*{20}{l}}
					{A_n^{i - 1} + 1,{\text{ }}a_{n,m}^{i - 1} = 0}, \\ 
					{1,{\text{ }}a_{n,m}^{i - 1} = 1}, 
			\end{array}} \right.
		\end{aligned}
	\end{equation}	
	where $a_{n,m}^{i - 1}$ denotes the association indicator at the $(i - 1)$-th global HFL iteration, i.e., if client $n$ is orchestrated by the edge server $m$, $a_{n,m}^{i - 1} = 1$; otherwise $a_{n,m}^{i - 1} = 0$.
	The clients with higher MS have a higher probability to be associated with the edge server, as their data samples are more informative to update the global model.
	Therefore, the average MS of the entire HFL system can be guaranteed to be relatively low, which benefits the fast convergence of HFL.
	
	It is worthwhile to point out that, for the unity of expression, we normalize the above three input variables in a same manner as:
	\begin{equation}\label{Fuzzy_nor}
		\begin{aligned}
			NV = \frac{V}{{MV}} \times 100\%,
		\end{aligned}
	\end{equation}
	where $V$ denotes the input variable, $MV$ indicates the maximum value of the input variables, and $NV$ is the normalized value of the input variable.
	Note that for different inputs, $V$ and $MV$ need to be replaced with corresponding values.
	
	\subsection{Fuzzy Output}
	\begin{figure}[t]
		\includegraphics[width=0.5\textwidth]{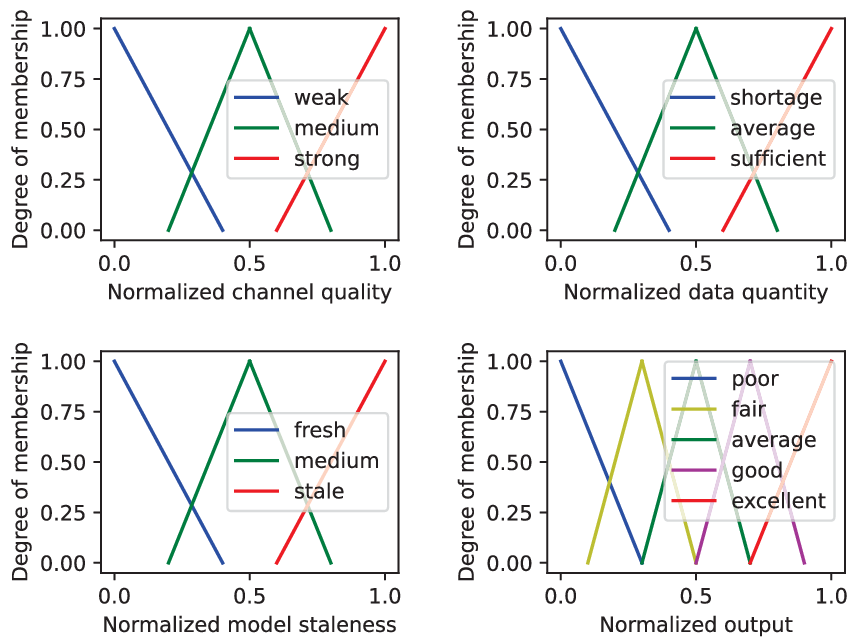}
		\centering
		\caption{Membership functions of fuzzy logic.}
		\label{FuzzyFunction}
	\end{figure}
	
	With fuzzy logic, the above three numerical inputs are transformed into fuzzy formats by a fuzzifier via the defined fuzzy membership functions in Fig. \ref{FuzzyFunction}.
	The transformed fuzzy input sets for these three criteria are given as 
	
	$CQ \in \left\{ {weak, medium, strong} \right\} $,
	
	$DQ \in \left\{ {shortage, average, sufficient} \right\} $,
	
	$MS \in \left\{ {fresh, medium, stale} \right\} $.

	Based on the three types of fuzzy inputs, we can obtain $27$ fuzzy rules in Table. \ref{FuzzyRule} to build a map between the fuzzy inputs and the fuzzy output.
	Define the fuzzy output set as:
	
	$Output \in \left\{ {poor, fair, average, good, excellent} \right\} $.
	
	The membership function of the fuzzy output for defuzzification is also defined in Fig. \ref{FuzzyFunction}.
	Note that in this paper, we utilize the Max-Min method \cite{wu2012vanet} to derive one rule from multi-criteria and combine multiple rules.
	To be specific, we choose the minimum value of the three membership functions as the degree of the fuzzy rule.
	When multiple rules correspond to the same output, we choose the rule with maximum degree.
	
		\begin{table}[t] 
		\footnotesize
		\begin{center}
			\caption{\protect\\\textsc{Fuzzy Logical Rules}}\vspace{+1em}
			\label{FuzzyRule}
			\begin{tabular}{|c|c|c|c|c|}
				\hline
				Rule & Channel quality & Data quantity & Model staleness & Output\\
				\hline
				1 & Strong &	Shortage &	Fresh &	Fair \\
				2 &	Strong &	Shortage &	Medium &	Average \\
				3 &	Strong &	Shortage &	Stale  &	Good \\
				4 &	Strong &	Average &	Fresh &	Average \\
				5 &	Strong &	Average	 & Medium &	Good \\
				6 &	Strong &	Average &	Stale &	Excellent \\
				7 &	Strong &	Sufficient &	Fresh &	Good \\
				8 &	Strong &	Sufficient &	Medium &	Excellent \\
				9 &	Strong &	Sufficient &	Stale &	Excellent \\
				10 &	Medium &	Shortage &	Fresh &	Poor \\
				11 &	Medium &	Shortage &	Medium &	Fair \\
				12 &	Medium &	Shortage &	Stale &	Average \\
				13 &	Medium &	Average &	Fresh &	Fair \\
				14 &	Medium &	Average	 & Medium &	Average \\
				15 &	Medium &	Average	 & Stale &	Good \\
				16 &	Medium &	Sufficient &	Fresh &	Average \\
				17 &	Medium &	Sufficient &	Medium &	Good \\
				18 &	Medium &	Sufficient &	Stale &	Excellent \\
				19 &	Weak &	Shortage &	Fresh &	Poor \\
				20 &	Weak &	Shortage &	Medium	 &Poor \\
				21 &	Weak &	Shortage &	Stale &	Fair \\
				22 &	Weak &	Average &	Fresh &	Poor \\
				23 &	Weak &	Average	 & Medium &	Fair \\
				24 &	Weak &	Average &	Stale &	Average \\
				25 &	Weak &	Sufficient &	Fresh &	Fair \\
				26 &	Weak &	Sufficient &	Medium	 &Average \\
				27 &	Weak &	Sufficient &	Stale &	Good \\
				
				\hline
			\end{tabular}
		\end{center}
	\end{table}
	
	After utilizing the Max-Min method described above, we can obtain multiple fuzzy outputs related to different fuzzy rules.
	Now, we need to defuzzify these fuzzy outputs into numerical formats based on the center of gravity (COG) method \cite{cha2022fuzzy}, which is expressed as
	\begin{equation}\label{COG}
		\begin{aligned}
			N{O^ * } = \frac{{\sum {NO \times f\left( {NO} \right)} }}{{\sum {f\left( {NO} \right)} }},
		\end{aligned}
	\end{equation}
	where $NO$ denotes the nomalized output, $f\left( {NO} \right)$ represents the corresponding membership function according to Fig. \ref{FuzzyFunction} and $N{O^ * }$ is the final nomalized output which indicates the clients' competency level to associate with the edge server.
	
	Then, each edge server sorts the clients located in its service coverage in a descending order according to their $N{O^ * }$, and selects the first $N_m$ clients to participate in edge model aggregation.
	Note that a client may be selected by more than one edge server.
	In this case, the client chooses the nearest edge server for association, and other edge servers will select a client to substitute in their queues.
	
	\section{Joint edge server scheduling and optimal resource allocation}
	
	\subsection{Problem Formulation}
	In this part, given the fuzzy-based client-edge association scheme, our objective is to achieve the total cost minimization in one global iteration of our proposed HFL system, while jointly optimizing edge server scheduling and resource allocation.
	Denote ${\lambda _t}, {\lambda _e} \in \left[0, 1\right]$ satisfing ${\lambda _t} + {\lambda _e} =1$ by the improtance weighting indicators related to the time consumption and energy consumption, respectively.
	Then, we can formulate the total cost minimization problem in the HFL system as
	\begin{subequations}\label{OrigPro}
		\begin{align}
			\mathop {\min }\limits_{{\bf{p}},{\bf{f}}, {\bf{z}}} \quad & {\lambda _t}T + {\lambda _e}E \label{} \\
			\st\ \quad  & p_n^{min}  \le  {p_n} \le p_n^{\max },\forall n \in {\cal N}_m,  \label{PowerCon}\\
			& f_n^{min}  \le {f_n} \le f_n^{\max },\forall n \in {\cal N}_m,  \label{FreqCon}\\
			& z_m \in \left\{ {0,1} \right\}, \forall m \in {\cal M},  \label{SelcCon}
		\end{align}
	\end{subequations}
	where constraints \eqref{PowerCon} and \eqref{FreqCon} limit the range of transmitting power and computing frequency of client $n$, respectively.
	Constraint \eqref{SelcCon} indicates that $z_m$ is a binary variable.
	
	Apparently, problem \eqref{OrigPro} is a non-convex mixed integer nonlinear programming (MINLP) problem, which is intractable to derive the optimal solutions directly.
	To efficiently address problem \eqref{OrigPro}, in this section, we propose to decompose it into two subproblems for edge server scheduling and resource allocation, and design the corresponding optimization algorithms to obtain the solutions iteratively.

	
	\subsection{Edge Server Scheduling}
	Given the resource allocation variables ${\bf{p}}$ and ${\bf{f}}$, problem \eqref{OrigPro} is reduced to schedule edge servers to minimize the total cost.
	Nevertheless, it is still intractable due to the existence of binary variable ${\bf{z}}$.
	To address this problem effectively, an auxiliary variable $\tilde z_m$ is introduced which is subject to the constraints of $ \tilde z_m = z_m$, $z_m \left(1 - \tilde z_m \right) = 0$, as well as $0 \le {{\tilde z}_m} \le 1, \forall m \in {\cal M}$.
	Note that these constraints guarantee $z_m$ equals to $ \tilde z_m$ only at $0$ or $1$.
	Besides, to deal with the Max expression in $T$, we define $U = \mathop {\max }\limits_{n \in {{\cal{N}}_m}} \left\{ {\tau _2} \left( {t_n^{cmp} + t_{n,m}^{com}}\right)  \right\} $ and $W = \mathop {\max }\limits_{m \in {{\cal{M}}}} \left\{ {z_m}\left( {T_m^{cloud} + U} \right)\right\}$.
	Then, the edge server scheduling subproblem can be transformed into a more tractable form, which is written as
	\begin{subequations}\label{Pro1}
		\begin{align}
			\mathop {\min }\limits_{{\bf{z}}, {\bf{\tilde z}}, U, W} \quad & {\lambda _t}W + {\lambda _e}\sum\limits_{m \in {\cal M}} {z_m}{\left( {E_m^{cloud} + E_{{{\cal N}_m}}^{edge}} \right)}\\
			\st\ \quad  & {t_n^{cmp} + t_{n,m}^{com}} \le U, \forall n \in {\cal N}_m, \forall m \in {\cal M}, \label{UCon1} \\
			& {z_m}\left( {T_m^{cloud} + U} \right) \le W, \forall m \in {\cal M}, \label{WCon1} \\
			& z_m - \tilde z_m = 0, \forall m \in {\cal M}, \label{Conz1} \\
			& z_m \left(1 - \tilde z_m \right) = 0, \forall m \in {\cal M}, \label{Conz2} \\
			& 0 \le {{\tilde z}_m} \le 1, \forall m \in {\cal M}. \label{Conz3} 
		\end{align}
	\end{subequations}
	
	In the following, a double-loop PDD-based algorithm \cite{PDD_TSP} is proposed to solve the problem \eqref{Pro1}. 
	Specifically, the PDD-based algorithm's inner loop solves the augmented Lagrangian (AL) problem, while the dual or penalty variables related to the constraints are updated at the outer loop.
	The AL form of problem \eqref{Pro1}, which is utilized to dualize and penalize the equality constraints \eqref{Conz1} and \eqref{Conz2}, can be presented as \cite{guo2018joint}
	\begin{subequations}\label{Pro2}
		\begin{align}
			\mathop {\min }\limits_{{\bf{z}}, {\bf{\tilde z}}, U, W} \quad 
			& \begin{array}{l}
			{\lambda _t}W + {\lambda _e}{\sum\limits_{m \in {\cal M}} {z_m}{\left( {E_m^{cloud} + E_{{{\cal N}_m}}^{edge}} \right)}} 
			\\ + \frac{1}{{2v}}\sum\limits_{m = 1}^M {{{\left[ {{z_m}\left( {1 - {{\tilde z}_m}} \right) + v{q_m}} \right]}^2}}\\
			+ \frac{1}{{2v}}\sum\limits_{m = 1}^M {{{\left( {{z_m} - {{\tilde z}_m} + v{{\tilde q}_m}} \right)}^2}} 
			\end{array} \label{ALPro}\\
			\st\ \quad  & {t_n^{cmp} + t_{n,m}^{com}} \le U, \forall n \in {\cal N}_m, \forall m \in {\cal M}, \label{UCon2} \\
			& {z_m}\left( {T_m^{cloud} + U} \right) \le W, \forall m \in {\cal M}, \label{WCon2} \\
			& 0 \le {{\tilde z}_m} \le 1, \forall m \in {\cal M}, \label{Conz3_2} 
		\end{align}
	\end{subequations}
	in which $v$ represents the non-negative penalty parameter, $q_m$ and $\tilde q_m$ denote the dual variables related to the equality constraints \eqref{Conz1} and \eqref{Conz2}, respectively.
	The AL problem \eqref{Pro2} is solved in the inner loop via problem decomposition, and the details are shown as follows.
	
	\subsubsection{Solution of ${{\tilde z}_m}$} 
	By fixing other variables, the unconstrained solution for ${{\tilde z}_m}$ can be derived directly by taking the first-order derivative towards the unconstrained function \eqref{ALPro} and enforcing the first-order optimality condition, which can be expressed as
	\begin{equation}\label{ti_z1}
		\begin{aligned}
			{\tilde z_m^u} = \frac{{z_m^2 + {q_m}{z_m}v + {z_m} + {{\tilde q}_m}v}}{{z_m^2 + 1}}.
		\end{aligned}
	\end{equation}
	Recalling that $0 \le {{\tilde z}_m} \le 1$, we can obtain the optimal solution of ${{\tilde z}_m}$ as
	\begin{equation}\label{ti_z2}
		\begin{aligned}
			\tilde z_m^* = \left\{ \begin{array}{l}
				\begin{array}{*{20}{c}}
					{0,}&{\tilde z_m^u \le 0,}
				\end{array}\\
				\begin{array}{*{20}{c}}
					{\tilde z_m^u,}&{0 < \tilde z_m^u < 1,}
				\end{array}\\
				\begin{array}{*{20}{c}}
					{1,} &{\tilde z_m^u \ge 1.}
				\end{array}
			\end{array} \right.
		\end{aligned}
	\end{equation}

	\subsubsection{Solution of ${{z}_m}$}
	To derive the optimal solution of $z_m$, we shall solve the subproblem with respect to $z_m$ by fixing other variables.
	The transformed problem is formulated as 
	\begin{subequations}\label{Proz}
		\begin{align}
			\mathop {\min }\limits_{{\bf{z}}} \quad
			& \begin{array}{l}  
				{\lambda _e}{\sum\limits_{m \in {\cal M}} {z_m}{\left( {E_m^{cloud} + E_{{{\cal N}_m}}^{edge}} \right)}} \\
				  + \frac{1}{{2v}}\sum\limits_{m = 1}^M {{{\left[ {{z_m}\left( {1 - {{\tilde z}_m}} \right) + v{q_m}} \right]}^2}}\\
				+ \frac{1}{{2v}}\sum\limits_{m = 1}^M {{{\left( {{z_m} - {{\tilde z}_m} + v{{\tilde q}_m}} \right)}^2}} 
			\end{array} \label{Prozz}\\
			\st\ \quad 
			& {z_m}\left( {T_m^{cloud} + U} \right) \le W, \forall m \in {\cal M}. \label{WConz}
		\end{align}
	\end{subequations}
	The following Lemma 1 provides the optimal solution of $z_m$.
	
	~\\
	{\noindent\bf{Lemma 1}} {\textit{The optimal solution $z_m^*$ can be given by
	\begin{equation}\label{Optz}
		\begin{aligned}
			z_m^* = \frac{{{I_m}v}}{{1 + {{\left( {1 - {{\tilde z}_m}} \right)}^2}}},
		\end{aligned}
	\end{equation}
	where ${I_m} = \frac{{{{\tilde z}_m}}}{v} - {\tilde q_m} - {q_m}\left( {1 - {{\tilde z}_m}} \right) - {\lambda _e}\left( {{E_m^{cloud} + E_{{{\cal N}_m}}^{edge}}} \right) - {\gamma _m}\left( {T_m^{cloud} + U} \right)$, and ${\gamma _m}$ is the Lagrange multiplier corresponding to the constraint \eqref{WConz}}}.
	
	~\\
	{\noindent\textit{Proof.}} It is easy to prove that the second-order derivative of  \eqref{Prozz} is always larger than zero, suggesting it is a convex function.
	Besides, considering the convex constraint \eqref{WConz}, the optimization problem \eqref{Proz} belongs to a convex problem.
	Hence, we can utilize the Lagrange dual method to obtain the optimal solution.
	The Lagrange function of problem \eqref{Proz} is formulated as 
	\begin{equation}\label{Lagz}
		\begin{aligned}
			{\cal L}\left( {\bf{z}},{\gamma _m} \right) &= {\lambda _e}{\sum\limits_{m \in {\cal M}} {z_m}{\left( {E_m^{cloud} + E_{{{\cal N}_m}}^{edge}} \right)}} \\
			&+ \frac{1}{{2v}}\sum\limits_{m = 1}^M {{{\left[ {{z_m}\left( {1 - {{\tilde z}_m}} \right) + v{q_m}} \right]}^2}}\\
			&+ \frac{1}{{2v}}\sum\limits_{m = 1}^M {{{\left( {{z_m} - {{\tilde z}_m} + v{{\tilde q}_m}} \right)}^2}} \\
			&+{\gamma _m}\left[ {z_m}\left( {T_m^{cloud} + U} \right) - W \right].
		\end{aligned}
	\end{equation}
	By using the first-order derivative of ${\cal L}\left( {\bf{z}},{\gamma _m} \right)$ with respect to $z_m$, we can derive 
	\begin{equation}\label{}
		\begin{aligned}
				\frac{{\partial {\cal L}\left( {{\bf{z}},{\gamma _m}} \right)}}{{\partial {z_m}}} &= {\lambda _e}\left( {E_m^{cloud} + E_{{N_m}}^{edge}} \right)\\
				&+ \frac{1}{v}\left[ {{z_m}\left( {1 - {{\tilde z}_m}} \right) + v{q_m}} \right]\left( {1 - {{\tilde z}_m}} \right)\\
				&+ \frac{1}{v}\left( {{z_m} - {{\tilde z}_m} + v{{\tilde q}_m}} \right)+ {\gamma _m}\left( {T_m^{cloud} + U} \right).
		\end{aligned}
	\end{equation}
	By setting $\frac{{\partial {\cal L}\left( {{\bf{z}},{\gamma _m}} \right)}}{{\partial {z_m}}} = 0$, the optimal solution of $z_m$ can be derived, which is shown in \eqref{Optz}.
	This completes the proof.

	\subsubsection{Solutions of $U$ and $W$}
	According to the definition of $U$ and $W$, the optimal solutions can be calculated as follows:
	\begin{equation}\label{U}
		\begin{aligned}
			U^* = \mathop {\max }\limits_{n \in {{\cal{N}}_m}}\left\{ \frac{{\tau _1}{{c_n}{D_n}}}{{{f_n}}} + \frac{d_n}{R_{n,m}}  \right\}, 
		\end{aligned}
	\end{equation}
	\begin{equation}\label{W}
		\begin{aligned}
			W^* = \mathop {\max }\limits_{m \in {{\cal{M}}}} \left\{z_m^* \left( \frac{d_m}{R_m} + {\tau_2}U^* \right) \right\}.
		\end{aligned}
	\end{equation}

	\subsubsection{Update of $q_m$ and $\tilde q_m$}
	After obtaining the optimal solutions for the AL problem \eqref{Pro2} in the inner loop of the proposed PDD-based algorithm, we update the dual variables $q_m$ and $\tilde q_m$ in the outer loop as follows:
	\begin{equation}\label{q_1}
		\begin{aligned}
			q_m^{l + 1} = q_m^l + \frac{1}{{{v^l}}}\left[ {z_m^*\left( {1 - \tilde z_m^*} \right)} \right],\forall m \in {\cal M}, 
		\end{aligned}
	\end{equation}
	\begin{equation}\label{q_2}
		\begin{aligned}
			\tilde q_m^{l + 1} = \tilde q_m^l + \frac{1}{{{v^l}}}\left[ {z_m^* - \tilde z_m^*} \right],\forall m \in {\cal M},
		\end{aligned}
	\end{equation}
	in which $l$ is the iteration index of the outer loop.
	
	In Algorithm \ref{algorithm1}, the PDD-based edge server scheduling algorithm is presented. 
	The double-loop Algorithm \ref{algorithm1} solves the AL problem and updates the dual variables in its inner loop and outer loop, respectively.
	It is revealed in \cite{PDD_TSP} that the PDD-based algorithm can converge to the stationary solutions of problem \eqref{Pro1}.
	
	\begin{algorithm}[t]
		\footnotesize
		\caption{PDD-based Edge Server Scheduling Algorithm.}
		\label{algorithm1}
		\begin{algorithmic}[1] 
			\STATE Initialize auxiliary and dual variables ${\bf{\tilde z}}$, $\bf{q}$ and ${\bf{\tilde q}}$;
			\STATE Set parameter $\varepsilon = 10^ {-4}$ and set the maximum iteration number as $L^{max} $.
			\STATE \textbf{for} outer loop index $l = 1$ to $L^{max}$ \textbf{do}
			\STATE \qquad \textbf{while} $F^l -F^{l -1 }  \ge \varepsilon $ \textbf{do}
			\STATE \qquad\qquad Obtain ${{\tilde z_m^*}}$ according to \eqref{ti_z1} and \eqref{ti_z2};
			\STATE \qquad\qquad Calculate the optimal edge server scheduling $z_m^*$ based on \eqref{Optz}; 
			\STATE \qquad\qquad Obtain $U^*$ and $ W^* $ from \eqref{U} and \eqref{W};
			\STATE \qquad \textbf{end while}
			\STATE \qquad Update $q_m^{l+1}$ and $\tilde q_m^{l+1} $ based on \eqref{q_1} and \eqref{q_2};
			
			\STATE \qquad Set $l = l + 1$.
			\STATE \textbf{end for}
		\end{algorithmic}
	\end{algorithm}
	
	\subsection{Resource Allocation}
	Given the scheduled edge servers, we then consider the resource allocation subproblem.
	Now, we can convert the original problem \eqref{OrigPro} to a total cost minimization problem with the variables of resource allocation, which is formulated as 
	\begin{subequations}\label{Pro_RA}
		\begin{align}
			\mathop {\min }\limits_{{\bf{p}},{\bf{f}}} \quad & {\lambda _t}T + {\lambda _e}E \label{Obj_RA} \\
			\st\ \quad  & p_n^{min}  \le  {p_n} \le p_n^{\max },\forall n \in {\cal N}_m,  \label{PowerCon_RA}\\
			& f_n^{min}  \le {f_n} \le f_n^{\max },\forall n \in {\cal N}_m. \label{FreqCon_RA}
		\end{align}
	\end{subequations}
	Owing to the coupled optimization variables in \eqref{Obj_RA}, it is still challenging to obtain the optimal solutions utilizing traditional optimization methods.
	Moreover, when considering the dynamic channel environment in our proposed system, it is too challenging for the traditional methods to obtain the solutions.
	In this case, the DRL method is an idea candidate, which has a low optimization complexity and can interact with the dynamic environment to achieve better performance.
	Since problem \eqref{Pro_RA} can be formulated as a Markov decision process (MDP) with the continuous action and state spaces, we adopt the DDPG-based approach to achieve optimal resource allocation in this paper.
	Note that in DDPG, all the associated clients are treated as one agent, as they all have the same optimization goal.
	
	Next, to have a better understanding towards DPPG, we first introduce the concept of MDP briefly.
	According to the theory of Markov process, MDP can be regarded as an optimal decision process with respect to a stochastic dynamic system.
	More specifically, given the observed state, a new decision can be made by the agent in MDP to transfer the state and calculate the probability of state transition.
	Then, the new decisions are made by the decision maker according to the new observed state.
	The process of MDP can be described mathematically as $\left\{{S, A, R, P} \right\}$, in which $S$ is the state of the agent, $A$ denotes the available action for the agent under $S$, $R$ denotes the achievable reward for transferring into the next state and $P$ represents the probability of state transition when the agent takes action $A$ under state $S$ \cite{RIS_DDPG}.
	The objective problem \eqref{Pro_RA} now can be transformed into the MDP problem, and the detailed process for modelling is presented as follows.
	
	\subsubsection{State space}
	We consider that both the dynamic channel information of the network as well as the agent state are included in the state space.
	Denote $S_j$ by the agent's state space at time slot $j$, which is described as ${S_j} = \left\{ {h_{n,m}^j,D_n,\forall n \in {{\cal N}_m},\forall m \in {{\cal M}}} \right\}$.
	The ${h_{n,m}^j}$ denotes the channel gain from client $n$ to edge server $m$ at $j$-th time slot, and $D_n$ represents the data size of client $n$.
	
	\subsubsection{Action space}
	Based on the observed state information, reasonable decisions are made by the agent for power and computational resource allocation.
	At time slot $j$, the agent's action space can be described as ${A_j} = \left\{ {p_n^j,f_n^j,\forall n \in {{\cal N}_m}} \right\}$.
	
	\subsubsection{Reward}
	By interacting with the environment with action $A_j$, the new state $S_{j+1}$ and the reward $R_j$ can be obtained.
	We adopt the straightforward principle that setting the negative total cost as the instantaneous reward, which is given by 
	\begin{equation}\label{}
		\begin{aligned}
			R_j = - ({\lambda _t}T + {\lambda _e}E).
		\end{aligned}
	\end{equation}
	The reason is that DDPG training aims to realize the reward maximization, while our formulated objective is to minimize the total cost.

	\begin{figure}[t]
		\includegraphics[width=0.49\textwidth]{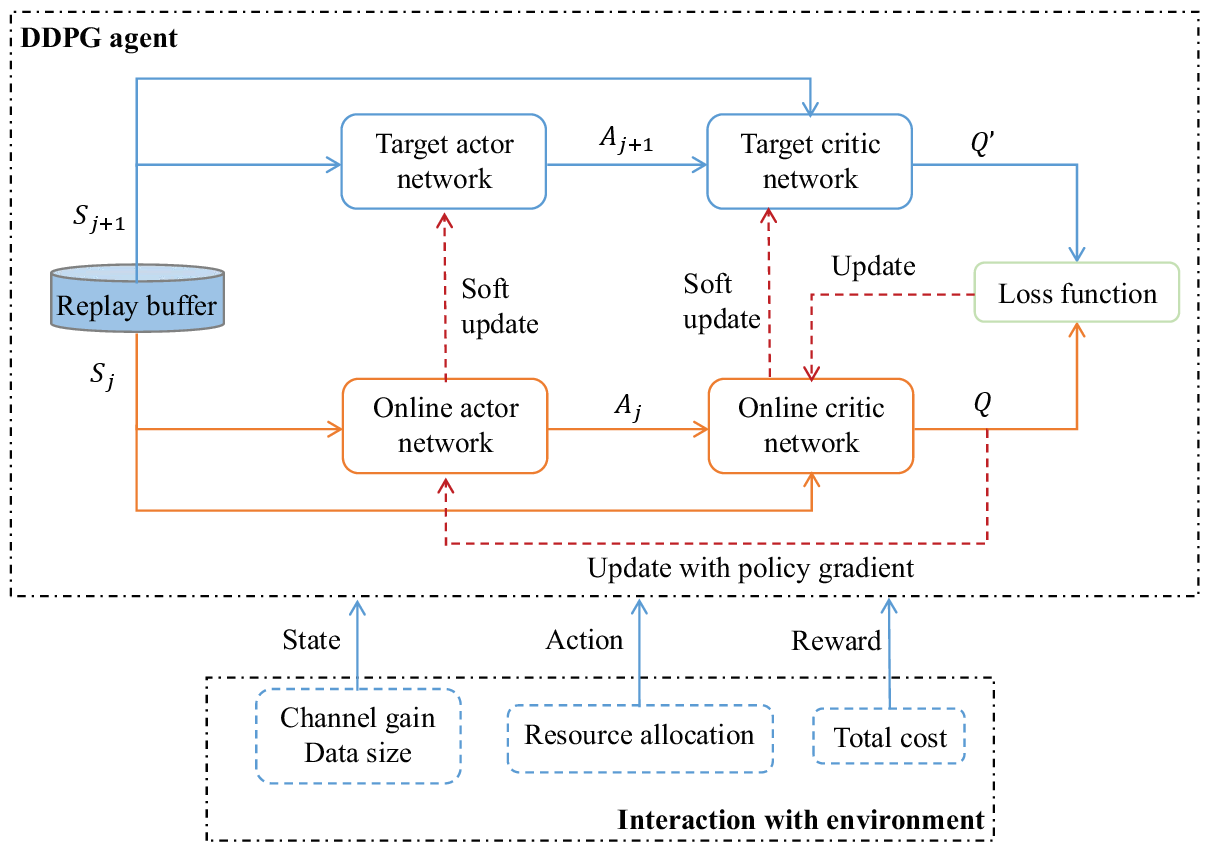}
		\centering
		\caption{The framework of DDPG training model.}
		\label{DDPG_model}
	\end{figure}
	
	In DDPG training model, there is an actor-critic framework, which is illustrated in Fig. \ref{DDPG_model}.
	Specifically, based on the current state, the actor network $\nu \left( {S|{{\bf{\theta }}^\nu }} \right)$ serves for action selection, while the critic network $Q\left( {S,A|{{\bf{\theta }}^Q}} \right)$ evaluates the action-state value.
	The ${{\bf{\theta }}^\nu }$ and ${{\bf{\theta }}^Q}$ are the corresponding weights related to neural networks.
	It is worthwhile to point out that each of the actor and critic networks has an online and a target network, so that the learning stability can be guaranteed and the overestimated issue in large-scale problems can be avoided.
	There is a similar structure between the target actor and critic network with the corresponding online networks, and they are defined as $\nu '\left( {S|{{\bf{\theta }}^{\nu '}}} \right)$ and $Q'\left( {S,A|{{\bf{\theta }}^{Q'}}} \right)$, respectively, in which ${{\bf{\theta }}^{\nu '}}$ and ${{\bf{\theta }}^{Q'}}$ are the related weights.
	Note that the experience replay is adopted during DDPG training.
	Particularly, at $j$-th time slot, the tuple $\left( {{S_j},{A_j},{R_j},{S_{j + 1}}} \right)$ is stored in the experience replay buffer.
	When reaching the maximum buffering threshold, the training of DDPG starts to update networks by sampling the mini-batch data from the experience replay buffer randomly.
	
	By minimizing the following loss function, the online critic network can be updated \cite{DDPG}:
	\begin{equation}\label{critic_net}
		\begin{aligned}
			L\left( {{{\bf{\theta }}^Q}} \right) = \frac{1}{{M'}}\sum\limits_j { {{{\left( {{y_j} - Q\left( {{S_j},{A_j}|{{\bf{\theta }}^Q}} \right)} \right)}^2}} },
		\end{aligned}
	\end{equation}
	where ${y_j} = {R_j} + \psi Q'\left( {{S_{j + 1}},\nu '\left( {{S_{j + 1}}|{{\bf{\theta }}^{\nu '}}} \right)|{{\bf{\theta }}^{Q'}}} \right)$, $M'$ denotes the mini-batch size and $\psi$ is the discount factor.
	Then, the online actor network is optimized by applying the sampled policy
	gradient ${\nabla _{{{\bf{\theta }}^\nu }}}Y$, which is given by
	\begin{equation}\label{actor_net}
		\begin{aligned}
			{\nabla _{{{\bf{\theta }}^\nu }}}Y = {{\mathbb{E}}_{{{\bf{\theta }}^\nu }}}\left[ {Q\left( {{S_j},\nu \left( {{S_j}|{{\bf{\theta }}^\nu }} \right)|{{\bf{\theta }}^Q}} \right)} \right],
		\end{aligned}
	\end{equation}
	where ${\mathbb{E}}\left(  \cdot  \right)$ presents the expectation operator.
	The target actor and critic networks are updated softly through following manner:
	\begin{equation}\label{target_update}
		\begin{aligned}
			{{\bf{\theta }}^{\nu '}} \leftarrow \zeta {{\bf{\theta }}^\nu } + \left( {1 - \zeta } \right){{\bf{\theta }}^{\nu '}},{{\bf{\theta }}^{Q'}} \leftarrow \zeta {{\bf{\theta }}^Q} + \left( {1 - \zeta } \right){{\bf{\theta }}^{Q'}},
		\end{aligned}
	\end{equation}
	where $\zeta$ controls the speed for updating.
	
	Algorithm \ref{algorithm2} summarizes the implementation of the DDPG-based resource allocation algorithm.

	\begin{algorithm}[t]
		\footnotesize
		\caption{DDPG-based Resource Allocation Algorithm.}
		\label{algorithm2}
		\begin{algorithmic}[1] 
			\STATE Randomly initialize the online actor network $\nu \left( {S|{{\bf{\theta }}^\nu }} \right)$ and critic network $Q\left( {S,A|{{\bf{\theta }}^Q}} \right)$ with weights ${{\bf{\theta }}^\nu }$ and ${{\bf{\theta }}^Q}$, respectively;
			\STATE Initialize the target actor network $\nu '$ and critic network $Q '$ with weights ${{\bf{\theta }}^{\nu '}} \leftarrow {{\bf{\theta }}^\nu },{{\bf{\theta }}^{Q'}} \leftarrow {{\bf{\theta }}^Q} $;
			\STATE Initialize the experience relay buffer and other network hyperparameters.
			\STATE \textbf{for} each episode \textbf{do}
			\STATE \qquad Initialize a random noise $n_j$ for action exploration;
			\STATE \qquad Receive the initial observed state $S_1$;
			\STATE \qquad \textbf{for} time slot $j = 1$ to $J$ \textbf{do}
			\STATE \qquad\qquad Select action $A_j = \nu \left( {S_j|{{\bf{\theta }}^\nu }} \right) + n_j$ based on the online 
			actor \\ \qquad\qquad network and the random exploration noise;
			
			\STATE \qquad\qquad Execute action $A_j$, obtain the reward $R_j$ and the next
			state \\ \qquad\qquad $S_{j+1}$;
			
			\STATE \qquad\qquad Store the tuple $\left( {{S_j},{A_j},{R_j},{S_{j + 1}}} \right)$ into the replay buffer;
			\STATE \qquad\qquad \textbf{if} reaching the maximum buffering threshold \textbf{then}
			\STATE \qquad\qquad\qquad Randomly sample $M'$ mini-batch data from the replay 
			\\ \qquad\qquad\qquad buffer;
			
			\STATE \qquad\qquad\qquad Update the online critic and actor network according to 
			\\ \qquad\qquad\qquad \eqref{critic_net} and \eqref{actor_net}, respectively;
			
			\STATE \qquad\qquad\qquad Update the target actor network and critic network based 
			\\ \qquad\qquad\qquad on \eqref{target_update}.
			
			\STATE \qquad\qquad \textbf{end if}
			\STATE \qquad \textbf{end for}
			\STATE \textbf{end for}
		\end{algorithmic}
	\end{algorithm}

	\begin{table}[t]
		\footnotesize
		\begin{center}
			\caption{\protect\\\textsc{Simulation Parameters}}\vspace{+1em}
			\label{Para}
			\begin{tabular}{c|c}
				\hline
				\hline
				Parameter & Value\\  \hline
				Bandwidth, $B $ & $1$ MHz \\
				Carrier frequency & $1$ GHz\\
				AWGN spectral density & $-174$ dBm/Hz\\
				Path loss exponent & $3.76$\\
				Transmit power limit, $p_n^{min}$, $p_n^{max}$ & $[0.01, 0.1] $  W\\
				CPU cycles for each sample, $\mu$ & $10^7$ \\
				Local capacitance coefficient, $\beta_n$ &  $10^{-28}$ \\
				Computation frequency limit, $f_n^{min}$, $f_n^{max}$ & $[1, 10]$ GHz\\
				Local model size, $d_n $ & $ 1$ Mbit\\
				Learning rate, $\eta$ & $0.01$ \\
				Weighting parameters, $\lambda_t$, $\lambda_e$ & 0.5 \\
				\hline
				\hline
			\end{tabular}
		\end{center}
	\end{table}
	
	\section{Simulations and analysis}
	In this part, abundant of simulations are conducted to verify the superior performance of our proposed schemes.
	A square area with side length of 500 meters is considered in our simulation, and the cloud server is located at its center.
	We considered 4 edge servers which are deployed at the midpoints of the lines connecting the four corners of the square to the cloud server, serving 64 randomly distributed clients.
	We set the number of associated clients for each edge server $N_m$ as 4 considering the designing complexity of the NOMA receiver.
	The HFL algorithm is used for the classification task using the MNIST dataset.
	We consider two data distribution scenarios in our simulation, including the independent and identically distributed (IID) data distribution and the non-IID data distribution.
	Specifically, for the former, the data distribution is the same among all clients, while in the non-IID scenario, different types of data labels are hold by different clients \cite{niid}.
	We present other simulation parameters in Table \ref{Para}.

	\subsection{Fuzzy Logic based Client-Edge Association}
	\begin{figure}[t]
		\includegraphics[width=0.4\textwidth]{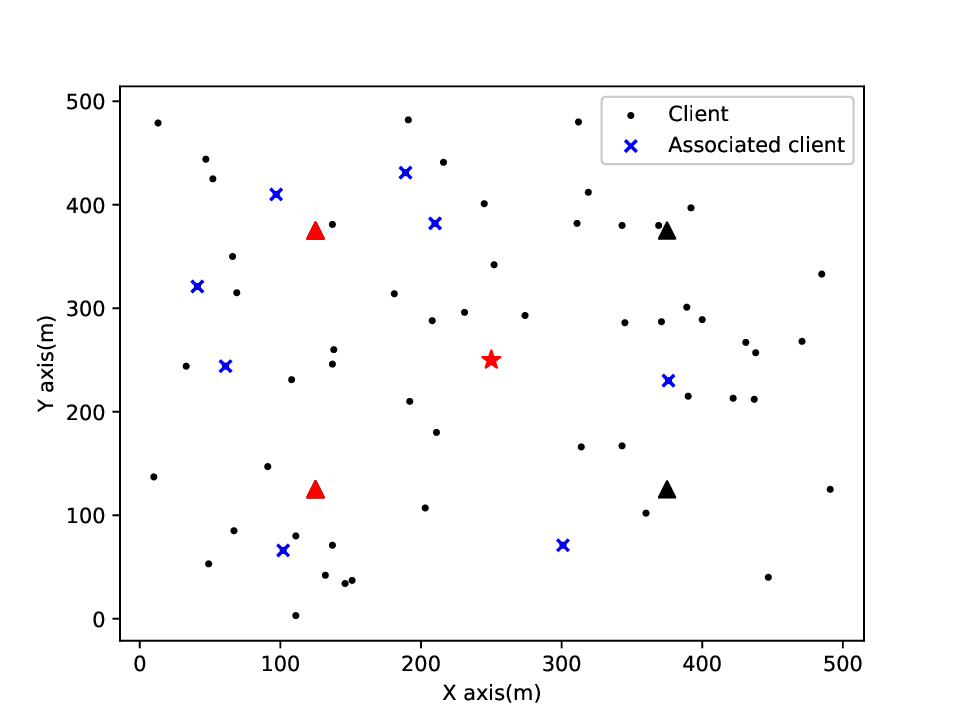}
		\centering
		\caption{Example of HFL system with fuzzy based association.}
		\label{Fuzzy_location}
	\end{figure}
	
	\begin{figure}[t]
		\includegraphics[width=0.4\textwidth]{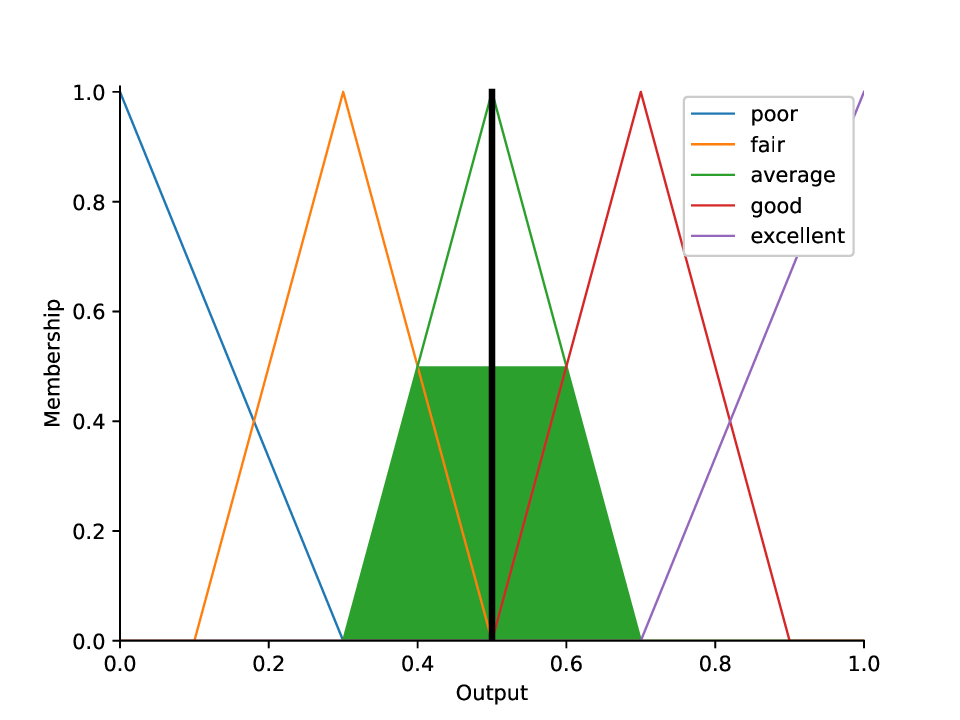}
		\centering
		\caption{Example of fuzzy output.}
		\label{Fuzzy_output}
	\end{figure}
	
	In Fig. \ref{Fuzzy_location}, we show an example of our proposed HFL system where the fuzzy logic is utilized to determine the client-edge association.
	The pentagram and triangles represent the cloud server and edge servers, respectively.
	Note that the red triangles in Fig. \ref{Fuzzy_location} denote the selected edge servers for cloud aggregation at this global round, while the black triangles are the unselected ones.
	It can be found that some clients which are not very close to the selected edge servers are still orchestrated due to their sufficient training data and their stale local models.
	This implies that our proposed scheme designs the client-edge association based on the comprehensive ability of clients rather than a single factor.
	Hence, the client heterogenerity in the HFL system can be balanced effectively.
	
	In Fig. \ref{Fuzzy_output}, an example of the fuzzy output with a simple input $(0.2, 0.5, 0.8)$ is presented.
	Note that the fuzzy input variables are normalized according to \eqref{Fuzzy_nor} for ease of analysis.
	Based on Fig. \ref{FuzzyFunction}, we find that the normalized fuzzy input of the client $(0.2, 0.5, 0.8)$ indicates that the channel quality is `weak', the data quantity is `average' and the local model is `stale'.
	Hence, according to rule 24 in Table \ref{FuzzyRule}, the corresponding fuzzy output is `average', and then the normalized output can be obtained from the x-axis value in Fig. \ref{Fuzzy_output}.
	For other more complicated cases with multiple fuzzy rules, the final normalized output is obtained through the COG method, as shown in \eqref{COG}.

	\subsection{Performance of HFL}
	To demonstrate the effectiveness of our proposed fuzzy based client-edge association (FCEA) scheme in improving HFL performance, we consider the following three benchmarking schemes, including the random client-edge association (RCEA) scheme, the greedy client-edge association (GCEA) scheme and the OMA scheme.
	Specifically, in RCEA, the association between clients and edge servers is designed in a random manner, while in GCEA, the client with the strongest channel gain builds association with the edge server, which reflects the single-criterion association.
	For the OMA scheme, the OMA transmission is considered between clients and the associated edge server, where the fuzzy based association is performed.

	\begin{figure}[t]
		\includegraphics[width=0.4\textwidth]{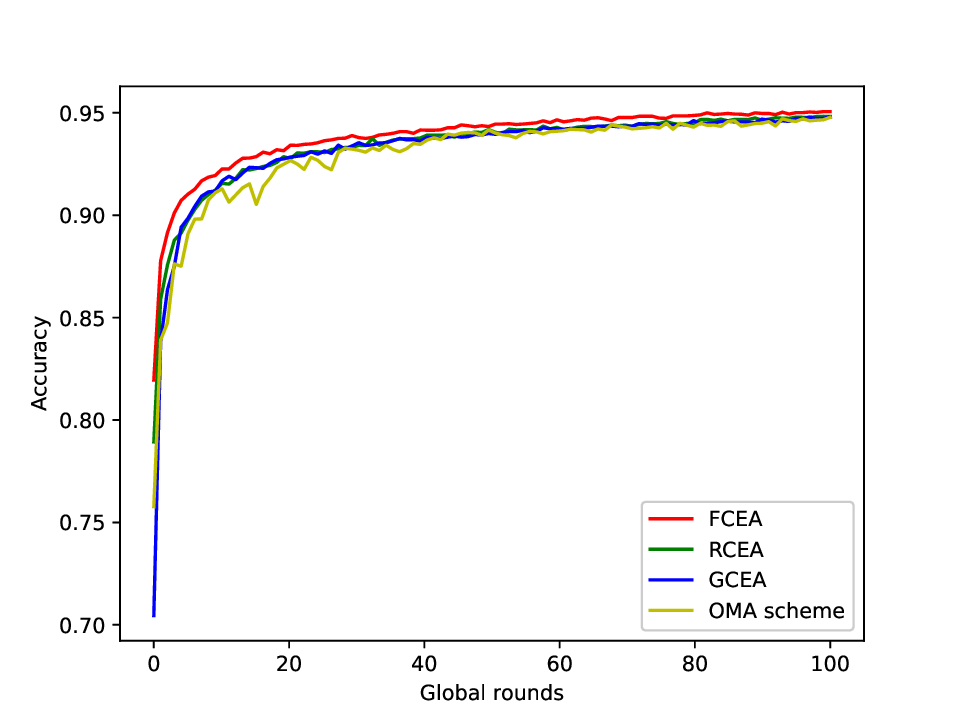}
		\centering
		\caption{Accuracy of HFL on IID data.}
		\label{HFL_acc_iid}
	\end{figure}
	
	\begin{figure}[t]
		\includegraphics[width=0.4\textwidth]{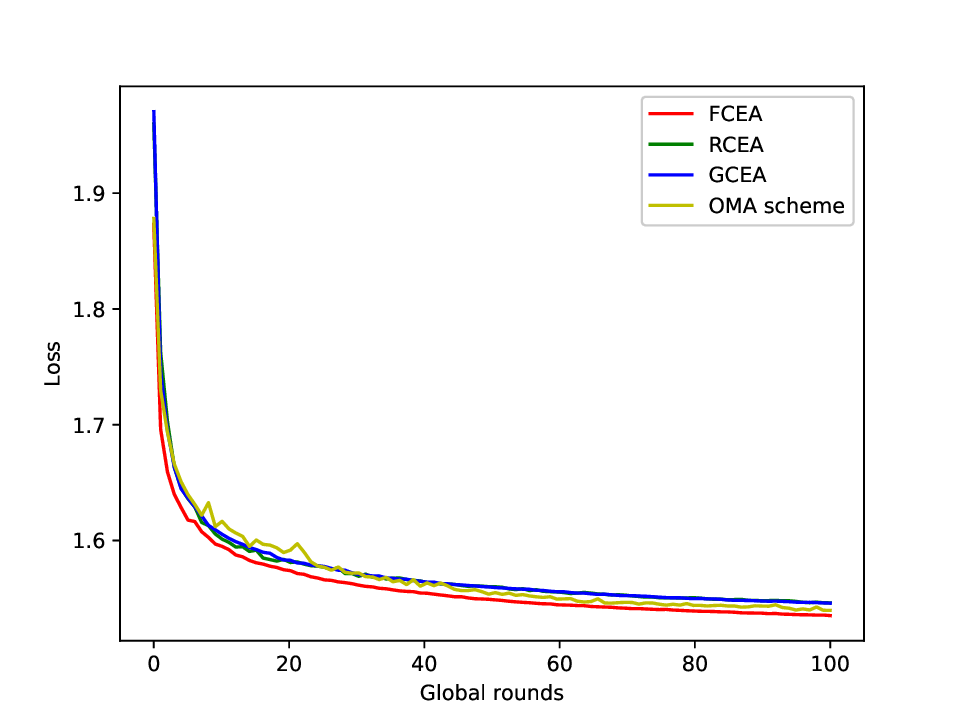}
		\centering
		\caption{Loss of HFL on IID data.}
		\label{HFL_loss_iid}
	\end{figure}

	Fig. \ref{HFL_acc_iid} illustrates the HFL accuracy on IID data.
	It can be observed that as HFL global round increases, before reaching the training convergence, there is a surge trend for the learning accuracy of all schemes.
	Moreover, the proposed FCEA scheme always achieves superior learning accuracy in comparison with other three benchmarks.
	This is because our scheme balances the client heterogenerity for client-edge association by accounting for three properties of clients.
	The global loss of HFL on IID data is presented in Fig. \ref{HFL_loss_iid}.
	As HFL global round grows, the HFL training loss decreases dramatically until convergence.
	It can also be observed that our proposed FCEA scheme can realize the lowest global HFL loss among the considered schemes.
	Therefore, it is validated that the proposed FCEA scheme can improve HFL performance.
	Besides, Fig. \ref{HFL_acc_iid} and Fig. \ref{HFL_loss_iid} show that the figures for the OMA scheme fluctuate severely before reaching convergence, which is owing to insufficient orchestrated clients at each round.
	Under this circumstance, more training rounds are required to achieve HFL convergence using enough data.
	At HFL convergence, the OMA scheme with fuzzy-based association can still achieve better performance in comparison with the RCEA and GCEA schemes.

	\begin{figure}[t]
		\includegraphics[width=0.4\textwidth]{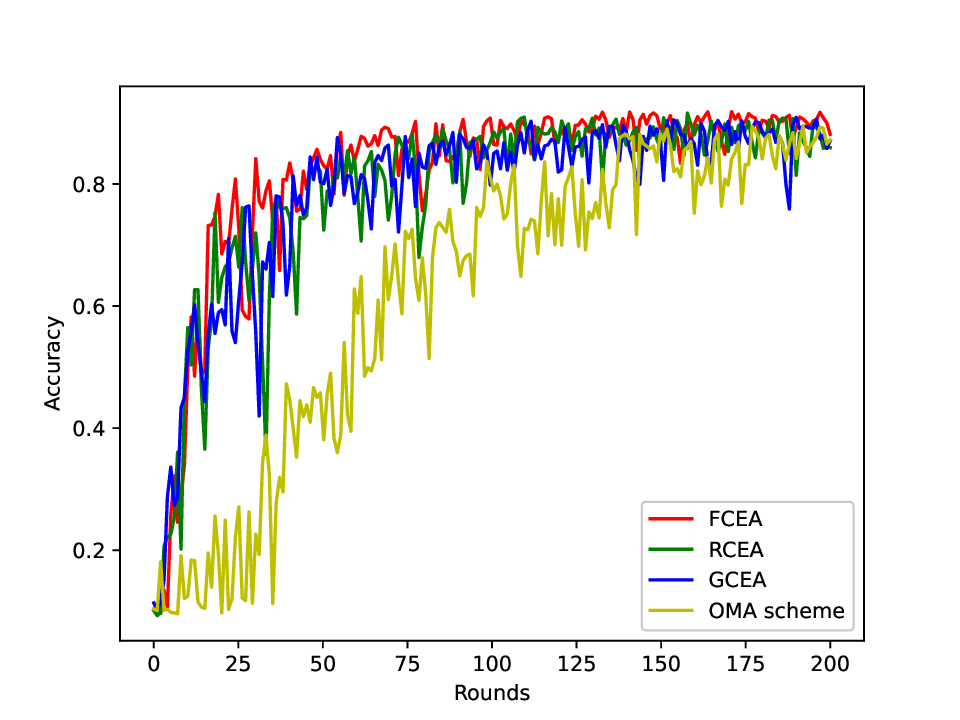}
		\centering
		\caption{Accuracy of HFL on non-IID data.}
		\label{HFL_acc_niid}
	\end{figure}
	
	\begin{figure}[t]
		\includegraphics[width=0.4\textwidth]{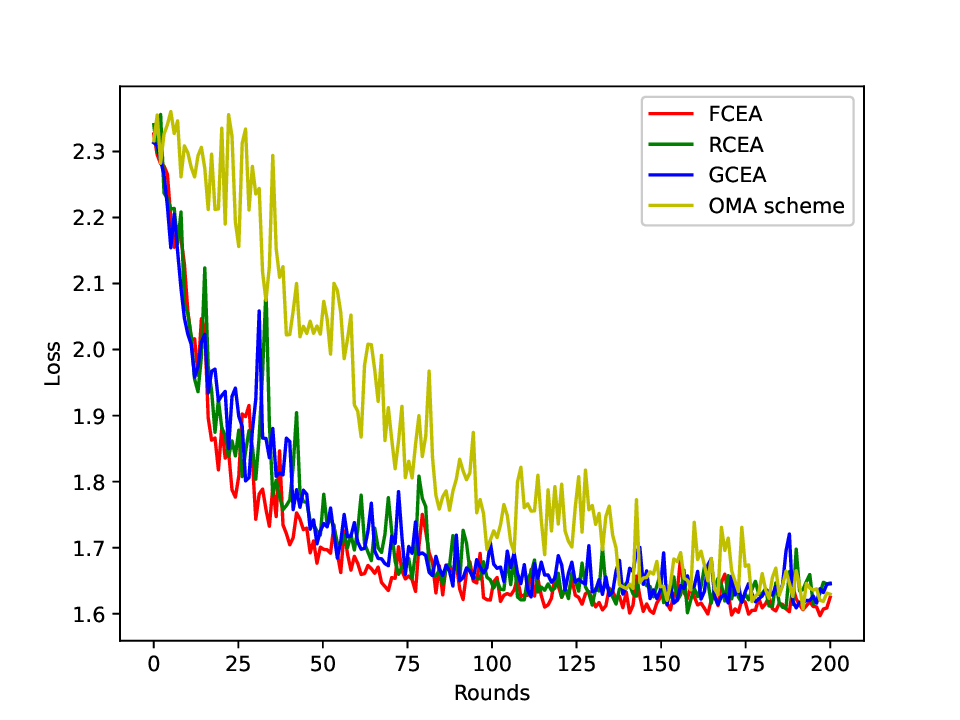}
		\centering
		\caption{Loss of HFL on non-IID data.}
		\label{HFL_loss_niid}
	\end{figure}
	
	Fig. \ref{HFL_acc_niid} and Fig. \ref{HFL_loss_niid} illustrate the HFL accuracy and global loss on non-IID data, respectively.
	Owing to the property of non-IID data, we can observe that both the HFL performance and robustness drop a lot comparing with the figures in Fig. \ref{HFL_acc_iid} and Fig. \ref{HFL_loss_iid}.
	Moreover, it requires more global rounds to achieve HFL convergence with non-IID data distribution.
	Fig. \ref{HFL_acc_niid} and Fig. \ref{HFL_loss_niid} also reveal that the proposed FCEA scheme is able to realize better HFL performance regarding training accuracy and global loss in comparison with other three schemes.
	This is because the client-edge association with multiple criteria is considered in our scheme, which is more beneficial for the improvement of HFL performance.
	We can also observe that the OMA scheme owns the worst HFL convergence performance because of the insufficient associated clients at each round.
	Thus, the superiority of NOMA-enabled HFL system can be demonstrated.

	\subsection{Performance of Average MS}
	\begin{figure}[t]
		\includegraphics[width=0.4\textwidth]{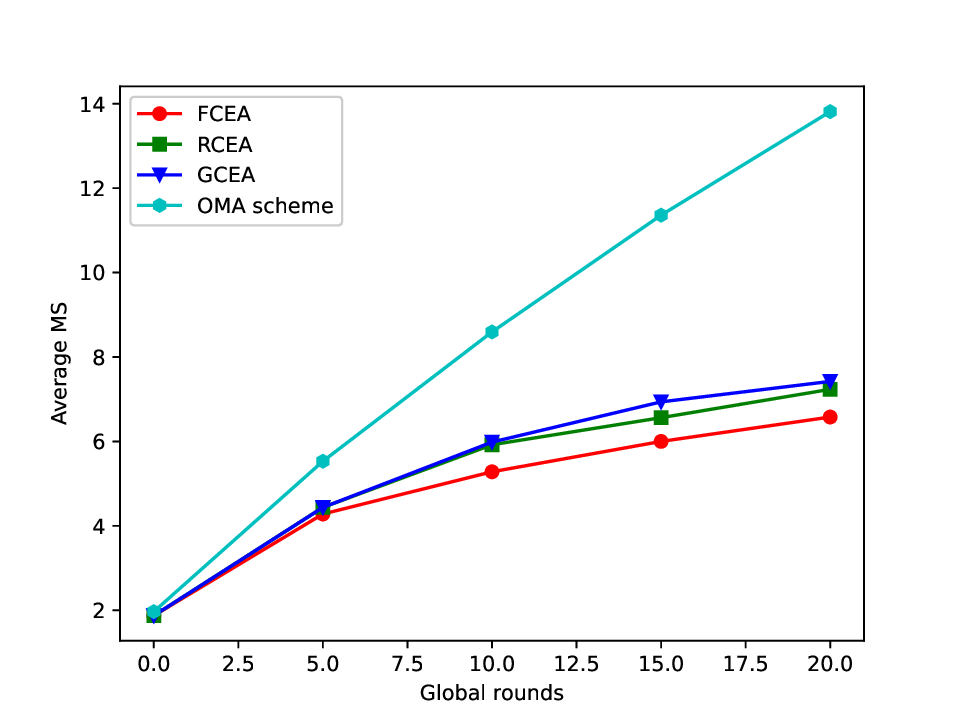}
		\centering
		\caption{Average MS of HFL system.}
		\label{HFL_age}
	\end{figure}
	
	Fig. \ref{HFL_age} illustrates the performance of average MS value which is defined in \eqref{AoU} versus global rounds over the above four schemes.
	It is obvious to find out that as HFL training round rises, the average MS figures of all schemes increase, while our proposed FCEA scheme achieves the lowest MS value in comparison with other schemes.
	This is because in our proposed FCEA scheme, the MS is regarded as one of the factors for client-edge association.
	Under this circumstance, the system average MS value can be guaranteed at a relatively low level, which is beneficial for the HFL convergence.
	Specifically, the low level average MS in HFL systems guarantees the local models to be fresh enough for model aggregation and more types of data can be utilized for HFL training especially in the scenarios with non-IID data distribution.
	In Fig. \ref{HFL_age}, we can also observe that the RCEA scheme and the GCEA scheme realize almost the same average MS value as HFL training.
	This is because in these two schemes, the MS is not considered in the design of client-edge association.
	Nevertheless, the average MS value of the OMA scheme increases more quickly in comparison with other three schemes because of the much less orchestrated clients at each HFL round.
	Under this circumstance, the average MS cannot be guaranteed even considering this factor in the design of association.

	\subsection{DDPG-based Resource Allocation}
	In this subsection, the performance of the DDPG-based resource allocation (DDPG-RA) algorithm is demonstrated given the FCEA scheme.
	For comparison, we consider three benchmarks which are also under the FCEA scheme, including the random resource allocation (RRA) scheme, the fixed power allocation (FPA) scheme \cite{xu2021adaptive} as well as the fixed computation allocation (FCA) scheme \cite{wen2022joint}.
	More specifically, in the RRA scheme, the power and computation resource are allocated randomly for clients, while the power and computation frequency are fixed in the FPA scheme and the FCA scheme, respectively.
	Note that the other variables in the benchmarks are optimized in the same way as the DDPG-RA algorithm to ensure the comparison fairness.
	
%
%
	
	\begin{figure}[t]
		\includegraphics[width=0.4\textwidth]{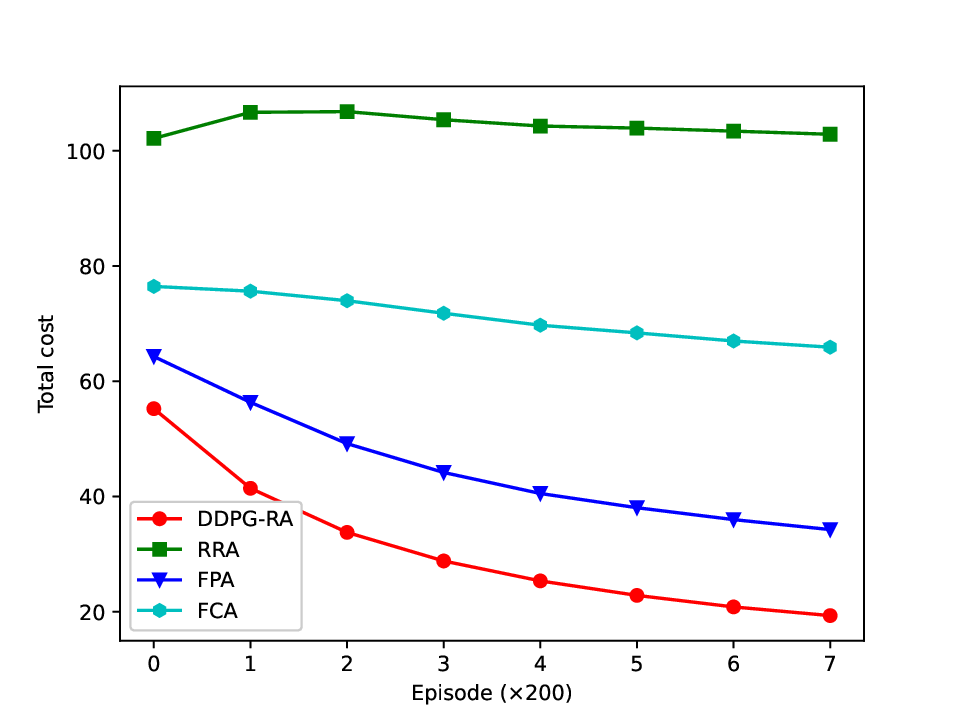}
		\centering
		\caption{Performance of total cost versus episode.}
		\label{Cost_episode}
	\end{figure}
	
	In Fig. \ref{Cost_episode}, the performance of average cost over four schemes as DDPG training is presented.
	As the training episode increases, the figures for schemes except the RRA scheme have the gradual decreasing trend, which verifies the effectiveness of efficient resource allocation to the total cost reduction.
	Particularly, for the RRA scheme, both power and computation resource are allocated randomly without considering optimization, hence leading to the fluctuation of its figure.
	For the other three schemes with the consideration of resource allocation optimization, it can also be found that our proposed DDPG-RA scheme realizes the lowest total cost compared with the FPA scheme and the FCA scheme.
	It validates the superiority of the joint optimization of the power and computational resource allocation.
	In the comparison of the FPA scheme and the FCA scheme, we find that the FPA scheme can achieve less total cost than the counterpart, which is determined by the weighting factors in the total cost definition.

	\begin{figure}[t]
		\includegraphics[width=0.4\textwidth]{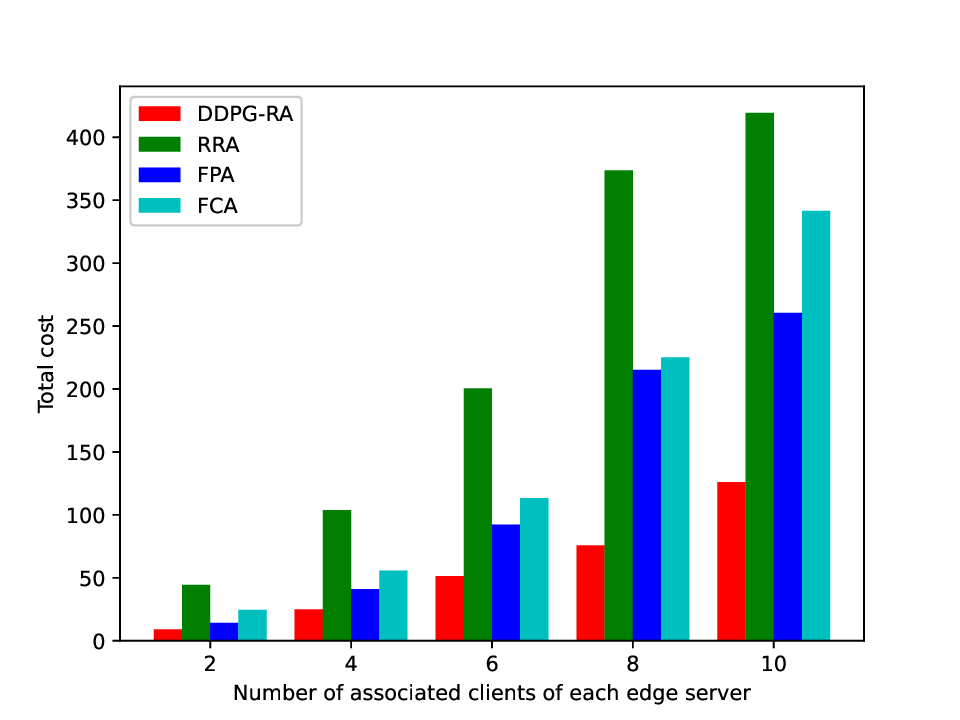}
		\centering
		\caption{Performance of total cost versus $N_m$.}
		\label{Cost_Nm_bar}
	\end{figure}
	
	\begin{figure}[t]
		\includegraphics[width=0.4\textwidth]{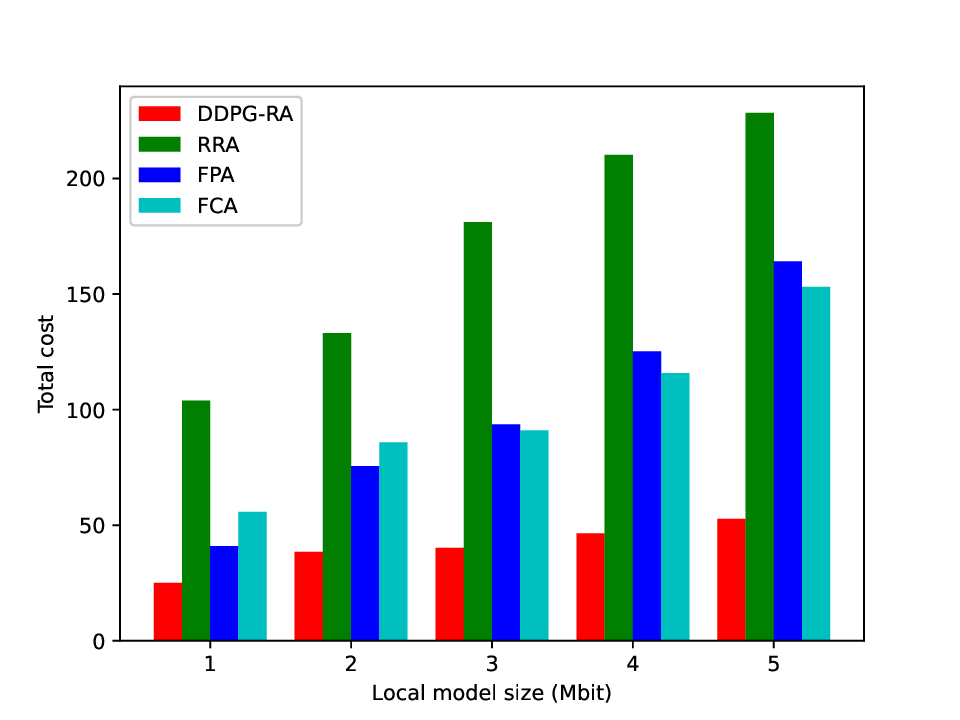}
		\centering
		\caption{Performance of total cost versus $d_n$.}
		\label{Cost_dn_bar}
	\end{figure}
	
	The performance of total cost versus different associated client numbers is shown in Fig. \ref{Cost_Nm_bar}.
	Apparently, the increasing $N_m$ results in the growth of total cost, which can be explained from following two aspects.
	On the one hand, with the growth of $N_m$, the total energy consumption of the HFL system increases definitely.
	On the other hand, increasing $N_m$ causes severe straggler issue for the synchronous edge model aggregation, which also increases the total time consumption in the HFL system.
	Thus, combining these two aspects, the total system cost increases with the growth of $N_m$.
	In Fig. \ref{Cost_Nm_bar}, we also find that our proposed DDPG-RA scheme always outperforms other schemes regarding the total cost minimization, which verifies the effectiveness of jointly optimizing transmit power and computational resources in the HFL system.
	Specifically, when $N_m = 4$, compared with the RRA, FPA and FCA schemes, our proposed DDPG-RA scheme is more efficient and fulfills up to $75.9\% $, $39\%$ and $55.1\% $ performance gain in reducing global cost, respectively.
	
	The performance of total cost versus the client $n$'s local model size $d_n$ is illustrated in Fig. \ref{Cost_dn_bar}. 
	The figure clearly shows that there is a rising trend for the total cost as $d_n$ increases.
	This is because the increasing $d_n$ results in the growth of wireless transmission time for the edge model aggregation, so that the corresponding energy consumption also increases.
	Compared with the benchmarks, our proposed algorithm always achieves the least total cost.
	In particular, take $d_n = 2$ Mbit as an example, the proposed DDPG-RA scheme realizes $71.1\% $, $49.1\%$ and $55.2\% $ performance gain in global cost reduction comparing with the RRA scheme, the FPA scheme and the FCA scheme, respectively.
	Moreover, it is interesting to find that as $d_n$ increases, the increase in total cost for the proposed DDPG-RA scheme is much less than the increase for the other three schemes.
	It indicates that the DDPG-RA scheme is potential in dealing with large local models.

	\section{Conclusion}
	In this paper, we studied the multi-criteria client orchestration and joint optimization of edge server scheduling and resource allocation in a NOMA enabled HFL system under semi-synchronous cloud model aggregation.
	The application of NOMA in HFL systems ensures the enhanced communication efficiency.
	We divided the system implementation into two phases, i.e., the client-edge phase and the edge-cloud phase, and formulated the corresponding time and energy cost, respectively.
	A fuzzy logic based client-edge association policy was first proposed to balance the client heterogenerity in multiple aspects.
	Then, we formulated the total cost minimization problem and decomposed it into edge scheduling and resource allocation two subproblems.
	The former subproblem was solved effectively through the PDD-based algorithm, and a closed-form solution of edge server scheduling was obtained.
	Considering the time-varying communication environment, we developed a DDPG-based algorithm to achieve efficient resource allocation among orchestrated clients at each edge server.
	Through extensive simulation results, we demonstrated that the proposed schemes own superior performance in improving HFL performance and reducing total cost.

	\bibliographystyle{IEEEtran}
	\bibliography{EEref}
	\vspace{0.5em}
\end{document}